\documentclass{article}

\usepackage{arxiv}

\usepackage[T1]{fontenc}    
\usepackage{url}            
\usepackage{booktabs}       
\usepackage{amsfonts}       
\usepackage{nicefrac}       
\usepackage{microtype}      
\usepackage{lipsum}
\usepackage{scalefnt}

\usepackage[latin1]{inputenc}
\usepackage{multirow}
\usepackage{lineno,hyperref}
\modulolinenumbers[5]
\usepackage{caption}
\usepackage{subcaption}
\usepackage{float}
\usepackage{todonotes}
\usepackage{textcomp}
\usepackage{amsmath}
\usepackage{eucal}
\usepackage{setspace}
\usepackage{lscape}

\DeclareMathOperator*{\argmax}{arg\,max}

\title{Automatic Chronic Degenerative Diseases Identification Using Enteric Nervous System Images}

\author{
  Gustavo Z.~Felipe\\
  Universidade Estadual de Maring\'a (UEM)\\
  Av. Colombo 5790, 87020-900\\
  Maring\'a, PR,  Brazil\\
  \texttt{gzf1996@gmail.com}\\
  \And
  Jacqueline N.~Zanoni\\
  Universidade Estadual de Maring\'a (UEM)\\
  Av. Colombo 5790, 87020-900\\
  Maring\'a, PR,  Brazil\\
  \texttt{zanonijn@uem.br}\\
  \And
  Camila C.~Sehaber-Sierakowski\\
  Universidade Estadual de Maring\'a (UEM)\\
  Av. Colombo 5790, 87020-900\\
  Maring\'a, PR,  Brazil\\
  \texttt{milla.sehaber@gmail.com}\\
  \And
  Gleison D. P.~Bossolani\\
  Universidade Estadual de Maring\'a (UEM)\\
  Av. Colombo 5790, 87020-900\\
  Maring\'a, PR,  Brazil\\
  \texttt{gleisonbossolani@gmail.com}\\
  \And
  Sara R. G.~Souza\\
  Universidade Estadual de Maring\'a (UEM)\\
  Av. Colombo 5790, 87020-900\\
  Maring\'a, PR,  Brazil\\
  \texttt{sara.raquel.gusman@gmail.com}\\
  \And
  Franklin C.~Flores\\
  Universidade Estadual de Maring\'a (UEM)\\
  Av. Colombo 5790, 87020-900\\
  Maring\'a, PR,  Brazil\\
  \texttt{fcflores@din.uem.br}\\
  \And
  Luiz E. S.~Oliveira\\
  Universidade Federal do Paran\'a (UFPR)\\
  Rua Cel. Francisco H. dos Santos 100, 81531-990\\
  Curitiba, PR, Brazil\\
  \texttt{luiz.oliveira@ufpr.br}\\
  \And
  Rodolfo M.~Pereira\\
  Instituto Federal do Paran\'a (IFPR)\\
  R. Humberto de A. C. Branco 1575, 83330-200\\
  Pinhais, PR, Brazil\\
  \texttt{rodolfomp123@gmail.com}\\
  \And
  Yandre M. G.~Costa\\
  Universidade Estadual de Maring\'a (UEM)\\
  Av. Colombo 5790, 87020-900\\
  Maring\'a, PR,  Brazil\\
  \texttt{yandre@din.uem.br}\\
}

\begin{document}
\maketitle

\begin{abstract}
Studies recently accomplished on the Enteric Nervous System have shown that chronic degenerative diseases affect the Enteric Glial Cells (EGC) and, thus, the development of recognition methods able to identify whether or not the EGC are affected by these type of diseases may be helpful in its diagnoses. In this work, we propose the use of pattern recognition and machine learning techniques to evaluate if a given animal EGC image was obtained from a healthy individual or one affect by a chronic degenerative disease. In the proposed approach, we have performed the classification task with handcrafted features and deep learning based techniques, also known as non-handcrafted features. The handcrafted features were obtained from the textural content of the ECG images using texture descriptors, such as the Local Binary Pattern (LBP). Moreover, the representation learning techniques employed in the approach are based on different Convolutional Neural Network (CNN) architectures, such as AlexNet and VGG16, with and without transfer learning. The complementarity between the handcrafted and non-handcrafted features was also evaluated with late fusion techniques. The datasets of EGC images used in the experiments, which are also contributions of this paper, are composed of three different chronic degenerative diseases: Cancer, Diabetes Mellitus, and Rheumatoid Arthritis. The experimental results, supported by statistical analysis, shown that the proposed approach can distinguish healthy cells from the sick ones with a recognition rate of 89.30\% (Rheumatoid Arthritis), 98.45\% (Cancer), and 95.13\% (Diabetes Mellitus), being achieved by combining classifiers obtained both feature scenarios.
\end{abstract}


\section{Introduction}
    The Enteric Nervous System (ENS), which the small intestine is dependent of, controls the digestive tract, coordinating different movement patterns such as: fast propulsion of content (peristalsis), mixing movements (segmentation), slow propulsion and retropulsion (expulsion of harmful substances associated with vomiting) \cite{furness06}.

Two networks, or neural plexus, compose the main components of the ENS: (1) an plexus located between the longitudinal and circular muscle layers, called myenteric (or Auerbach's) plexus; and (2) the submucosal (or Meissner's) plexus, located in the submucosa. The myenteric plexus controls the gastrointestinal movements. While the submucosal plexus, overall, controls the gastrointestinal secretion and the local blood flow \cite{furness06}. The Enteric Glial Cells (EGC) are another type of cells that can be found in the ENS as well. These ones play a vital role in the homeostasis of the gastrointestinal tract (GIT) functions \cite{Sharkey2015}.

Formerly, it was thought that the EGC worked only as a structural support to the neurons. But decisive studies carried out more recently \cite{Sharkey2012,Sharkey2015} verified that these cells also have other functions, contributing significantly to the neuronal maintenance, survival and function. In neurodegenerative processes, these cells play a role in the neuronal reconstruction, increasing their expression~\cite{ruhl11}. 

The immunostaining, i.e. an antibody-based method to detect a specific protein in a sample, of the S100 protein is used to identify the EGC. The S100, is a calcium binding protein located in the cytoplasm and/or nucleus of nerve and non-nerve tissues, can be expressed exclusively in the EGC. This one regulates the cytoskeleton's structure and function, as well as the calcium homeostasis in EGC's cytoplasm. It also presents neurotrophic properties that play a neuroprotective function~\cite{degiorgio12}.

The study of ENS cells is usually approached in pre-clinical research, aiming to experiment new methodologies and techniques in animals to be, later on, employed in humans. Resulting in the non-exposure of patients to the risk of death or permanent disability. Several projects involving the ENS have been developed by researchers in the field of neurogastroenterology \cite{Piovezana2019,Panizzon2019,Panizzon2016}. In these works, the enteric neurons and EGC are studied in order to understand the impact suffered by these cells on different diseases, as well as analyze the performance of different treatments for them. 

Usually, the enteric neurons and the EGC are preferable in such studies, because these kind of cells are all heavily affected by a considerable portion of chronic degenerative diseases. Thus, different studies may be taken on different diseases by evaluating a single kind of image sample.

Considering that the disease affects the EGC in shape and quantity, to ascertain the healthiness of a target animal, the researcher performed morphometric and quantitative analyses. These can be declared as exhaustive and time-consuming, once the lack of automation in the overall process, makes it extraordinarily manual and repetitive. It is reaffirming the relevance of developing computational models that execute such tasks automatically and with efficiency.

With that in mind, this work aimed to develop an approach for the automatic identification of chronic degenerative diseases in EGC animal images. Being capable of categorizing if an image sample from the ENS, evidencing the EGC, was obtained by a healthy or a sick animal. 

The proposed method aims to classify the images based on the extraction of handcrafted and non-handcrafted (automated learned) features. The handcrafted features are the ones here extracted by texture descriptors, while the non-handcrafted features are obtained with the use of Convolutional Neural Networks (CNNs). We have also investigated their complementarity by combining the resulting classifiers from both scenarios through classifiers' combination techniques. As far as we know, this is the first work to deal with identifying chronic degenerative diseases on ENS images.

To experimentally evaluate the proposed approach, we have created three datasets with EGC images of rats affected by different chronic degenerative diseases: Diabetes Mellitus, Cancer (Walkers Tumor-256), and Rheumatoid Arthritis. Each dataset is composed of image samples, collected by the evidence of EGC from the myenteric plexus from control (healthy) animals and from animals that presented the target disease. The datasets are freely available for download and can also be considered as a contribution to this work.

By achieving this goal, we look forward to giving the ENS researchers a texture-based automatic alternative to perform the EGC image analysis and foment new computer science research with the ENS, considering its great unexplored potential. It is worth mentioning that this one may be expanded to automatically detect diseases in human histopathological and radiographic images, aiming to reduce the possible subjectivity in their analysis. 

The reminding of this work is organized as follows: Section \ref{sec:methodology} presents the approach proposed in this work, describing the pre-processing, feature extraction, classification and combination phases. Section \ref{sec:database} presents the three degenerative diseases datasets proposed in this work. Section \ref{sec:results} presents the experimental analysis of the proposed approach, which is subdivided in exploratory investigation, parameters, configurations and results. In Section \ref{sec:discussions} we present the analysis and discussion regarding the obtained results and, finally, in Section \ref{sec:conclusions} we describe the concluding remarks and future works.


\section{Proposed Approach}
    \label{sec:methodology}

\begin{figure}[htpb!]
    \centering 
    \includegraphics[width=0.9\textwidth, keepaspectratio]{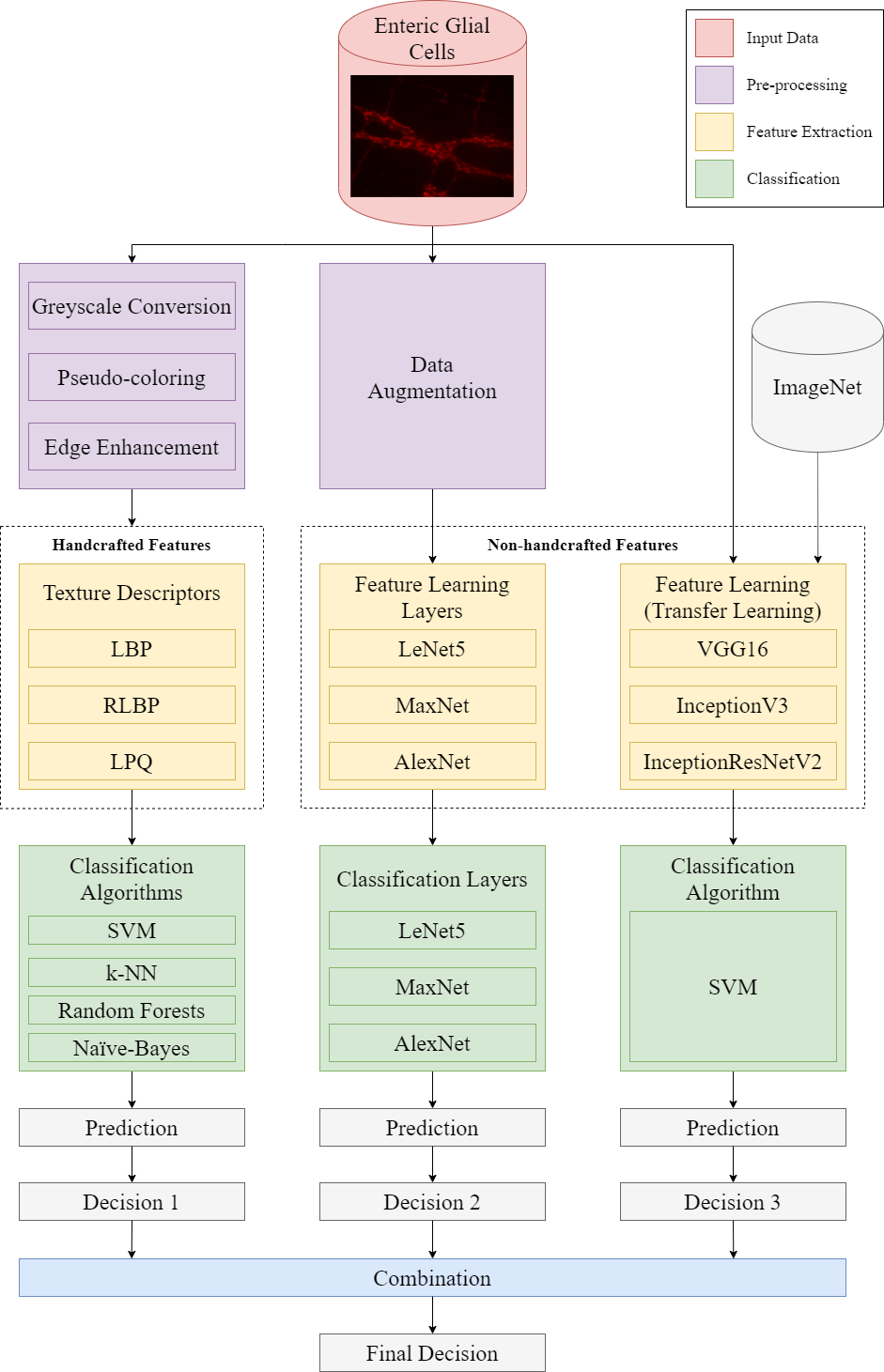}
    \caption{Representation of the approach used in this work, in which handcrafted and non-handcrafted features were approached.}
    \label{fig:overview-met}
\end{figure}

By analyzing the content of the images investigated in this work, we can observe that texture is one of the primary visual content to be explored. In this way, we decided to organize our proposed approach mainly based on different strategies aimed at describing the textural content. Besides that, methods that can also cooperate to  each other in the perspective of the combination of classifiers or representations. In this vein, we create descriptors founded both on the so-called handcrafted and non-handcrafted scenarios.  An overview of the proposed approach can be seen in Figure \ref{fig:overview-met}.

The handcrafted features correspond to features manually extracted, aiming to find the best representation of the addressed data through a process also known as feature engineering. These kinds of features were extracted in this work considering some widely and successfully used texture operators available in the literature. In total, three texture descriptors were broached: Local Binary Pattern (LBP)~\cite{lbp}, Robust Local Binary Pattern (RLBP)~\cite{zhao2013completed} and Local Phase Quantization (LPQ)~\cite{lpq}. 

The handcrafted features are then used as input to well-known classification algorithms such as Support Vector Machines (SVM), Random Forests (RF), k-Nearest Neighbors (k-NN) and Naive-Bayes (NB) classification algorithms. 

As aforementioned, the second scenario studied here extracts non-handcrafted features, i.e., features automatically extracted from the images, through feature learning techniques. In this work, we used three well-established Convolutional Neural Network (CNN) architectures to obtain this kind of feature: LeNet5 \cite{lenet5-89}, AlexNet \cite{alexnet12} and MaxNet \cite{roecker18}.

The concept of Transfer Learning was also experimented using pre-trained CNNs to extract features from the image samples. Thus, the CNN architectures VGG16 \cite{vgg16}, InceptionV3 \cite{inceptionv3} and InceptionResNetV2 \cite{inceptionresnet} were used. The feature learning layers of such models had their weights trained in the ImageNet dataset \cite{imagenet}. It is worth mentioning that the chi-square test ($X^2$) was employed as a feature selection method, aiming to reduce the number of features extracted from these CNN architectures. Differing from the traditional use of transfer learning, instead of redesigning the CNN model's classification layers, in this work, we classified the resulting features using the SVM algorithm. 

The estimation of probabilities generated from the classification experiments performed was then used to combine the resulting classifiers. Thus, the sum, product, and max classification rules, proposed by Kittler et al.~\cite{kittler1998}, were applied aiming to take advantage of a possible complementarity between classifiers generated by the use of different features and techniques.

It is worth mentioning that the experiments used the Stratified K-Fold Cross-Validation technique to divide the dataset to keep the existing proportions of the problem's classes. In this work, the $k$ value used was set to ten. More details about the concepts introduced here may be found in the following subsections.

\subsection{Handcrafted Modeling}
    This Section describes the handcrafted modeling, which follows the Pattern Recognition framework to classify features extracted through feature engineering. For simplicity we refer them as handcrafted features. 

    \subsubsection{Pre-Processing}
\begin{figure}[htpb!]
\centering
    \begin{subfigure}[h]{0.35\textwidth}
        \includegraphics[width=\textwidth, keepaspectratio]{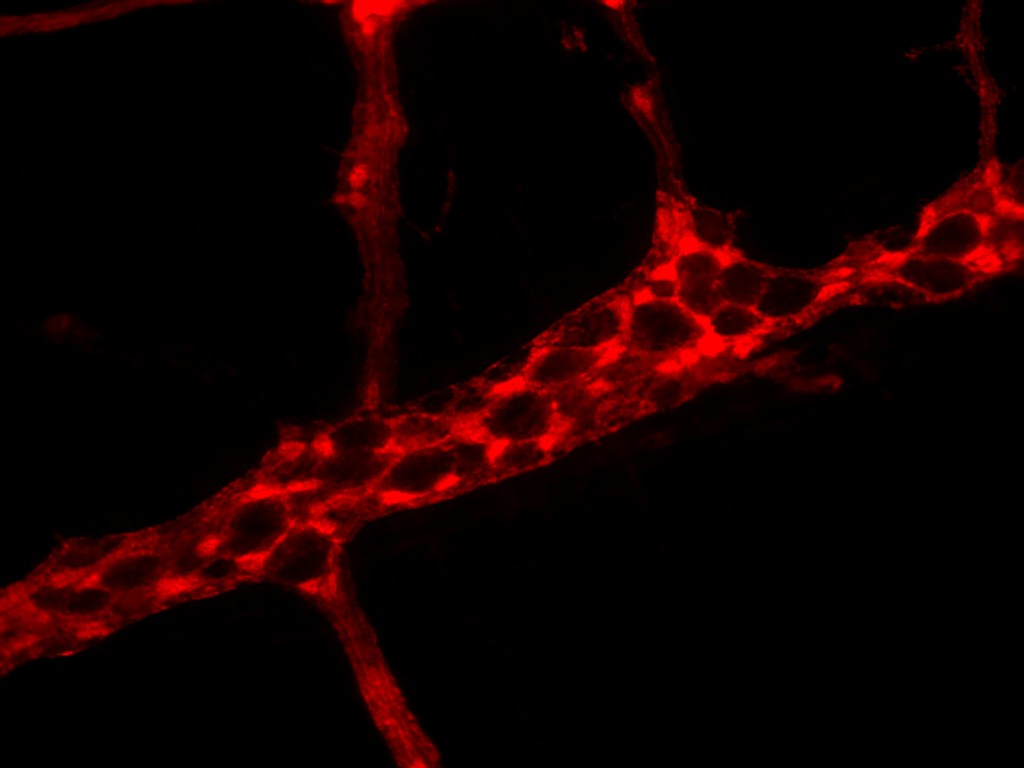}
    \end{subfigure}
    \begin{subfigure}[h]{0.35\textwidth}
        \includegraphics[width=\textwidth, keepaspectratio]{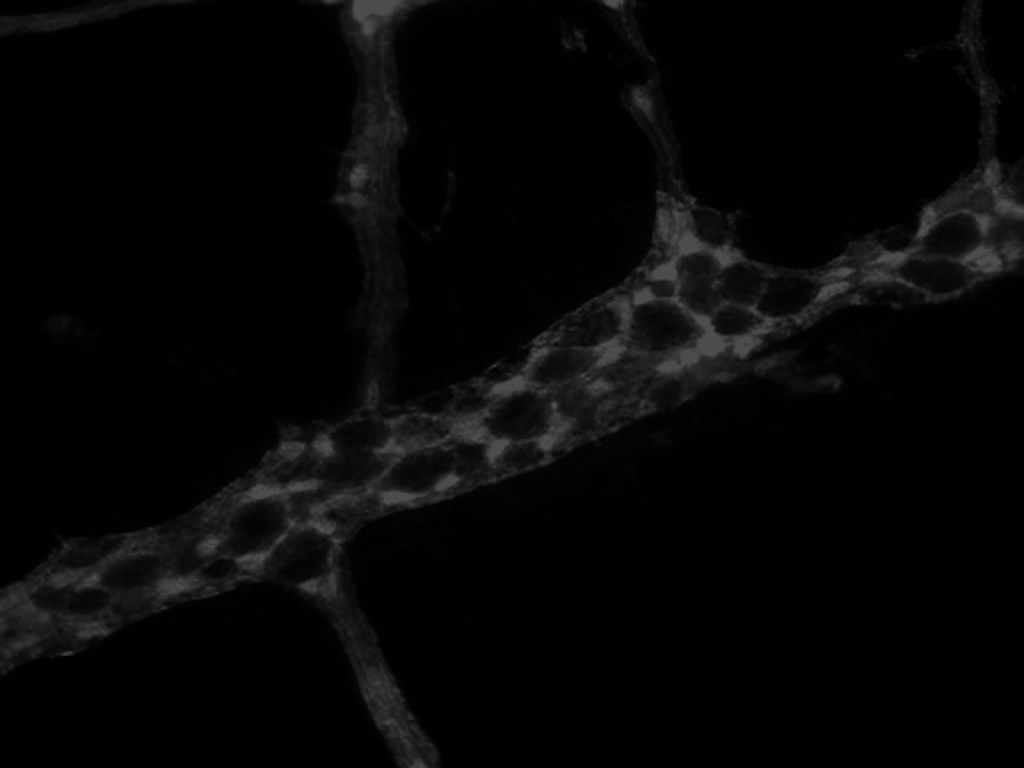}
    \end{subfigure}
    \caption{Digital image sample of the AIA dataset (left) 
    converted to the grayscale (right).}
    \label{fig:gliaCellGS}
\end{figure}

        In this work, some pre-processing techniques were explored, considering two main goals. The first one performs variations in the image samples' coloration. To achieve it, firstly, the image samples had their colors omitted by converting them to grayscale. Motivated by the fact that all image samples have a red tonality, implying that the two other channels of the RBG color system (blue and green) may not have any influence when extracting the handcrafted features. Figure \ref{fig:gliaCellGS} presents a comparison between an image sample in its original coloration and the same one after the conversion to grayscale.
        
\begin{figure}[htpb!]
\centering
    \begin{subfigure}[h]{0.35\textwidth}
        \includegraphics[width=\textwidth, keepaspectratio]{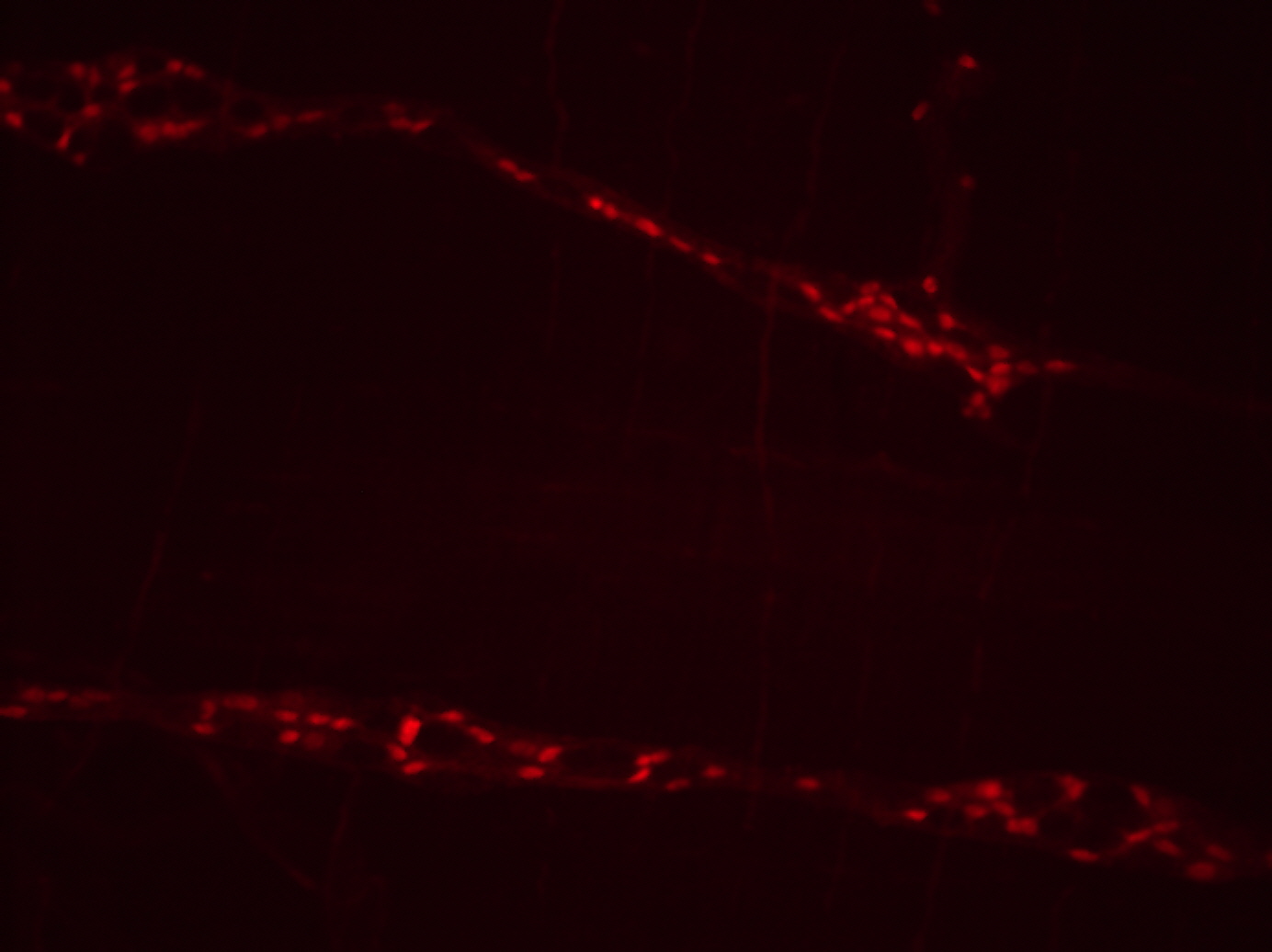}
        \caption{RGB}
    \end{subfigure}
    \begin{subfigure}[h]{0.35\textwidth}
        \includegraphics[width=\textwidth, keepaspectratio]{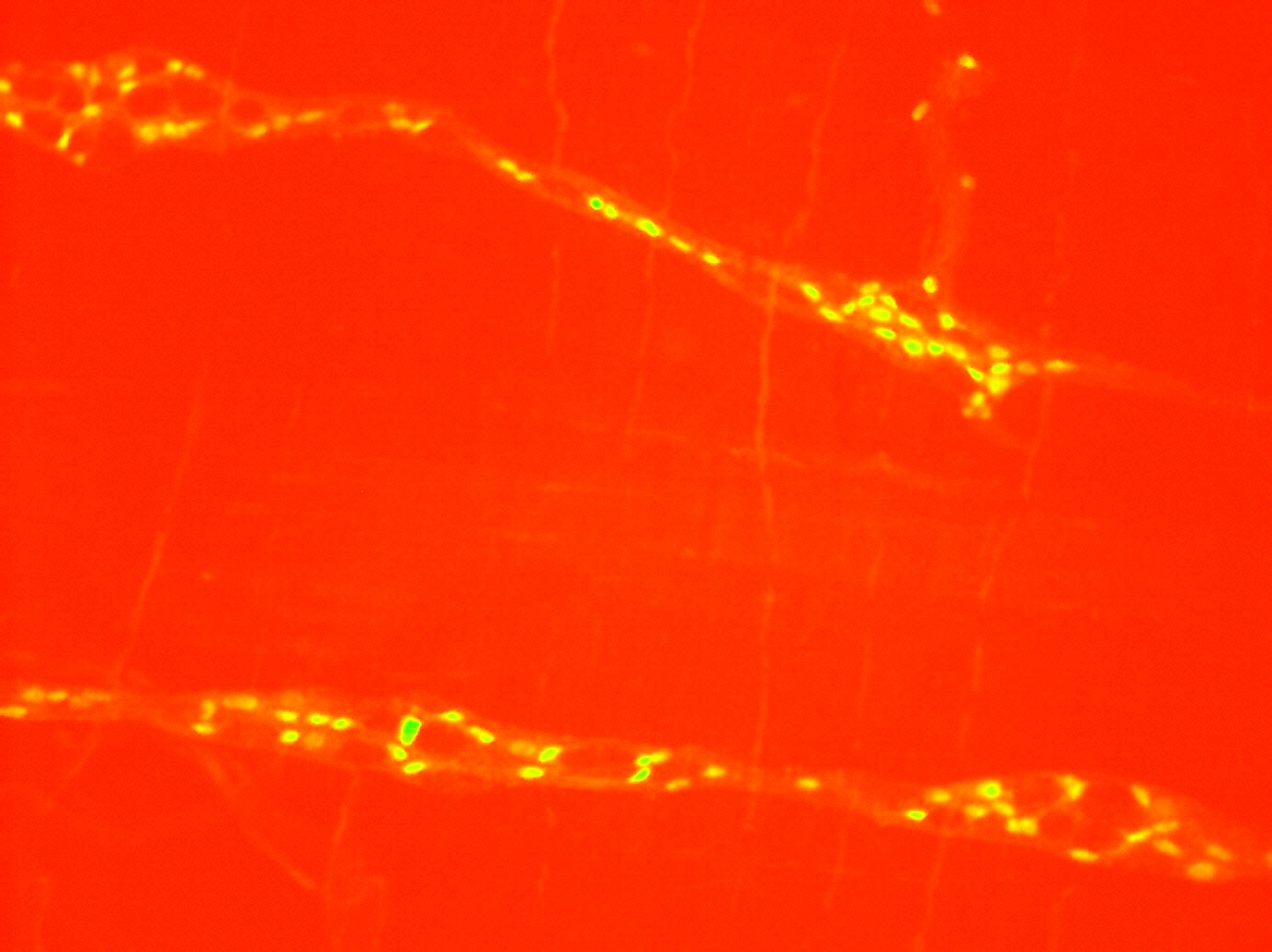}
        \caption{Pseudo-colored}
    \end{subfigure}
    \hfill
    \begin{subfigure}[h]{\textwidth}
        \centering
        \includegraphics[width=0.5\textwidth, keepaspectratio]{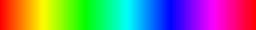}
        \caption{HSV Color Map}
    \end{subfigure}

    \caption{Digital image sample of the D disease group (a) and the same one pseudo-colored (b) using the HSV color map (c).}
    \label{fig:gliaCellHSV}
\end{figure}

Then, in a second approach, the image samples were pseudo-colored to create a new color pattern to highlight the EGC. The pseudo-coloring is a technique that tries to color a grayscale image sample. This is commonly achieved by mapping a single grayscale value to an RGB value, which is usually referred to as color maps. Since the image samples used in this work were originally in the RGB color system, it was necessary to convert them to the grayscale to apply such a technique. A representation of a pseudo-coloring may be observed in Figure \ref{fig:gliaCellHSV}.
    
It is worth mentioning that these kinds of operations do not affect the image samples' existing texture. 

Most of the captured EGC image samples have a lack of sharpness in their edges/shapes. These images result from the immunohistochemical reaction for a protein expressed exclusively in EGC, the S10 protein. This protein can be irregularly distributed in the cell, which can form irregular outlines and, associated with the low resolution of the microscopy used, often generates images with a blurred aspect.
        
\begin{figure}[htpb!]
\centering
    \begin{subfigure}[h]{0.35\textwidth}
        \includegraphics[width=\textwidth, keepaspectratio]{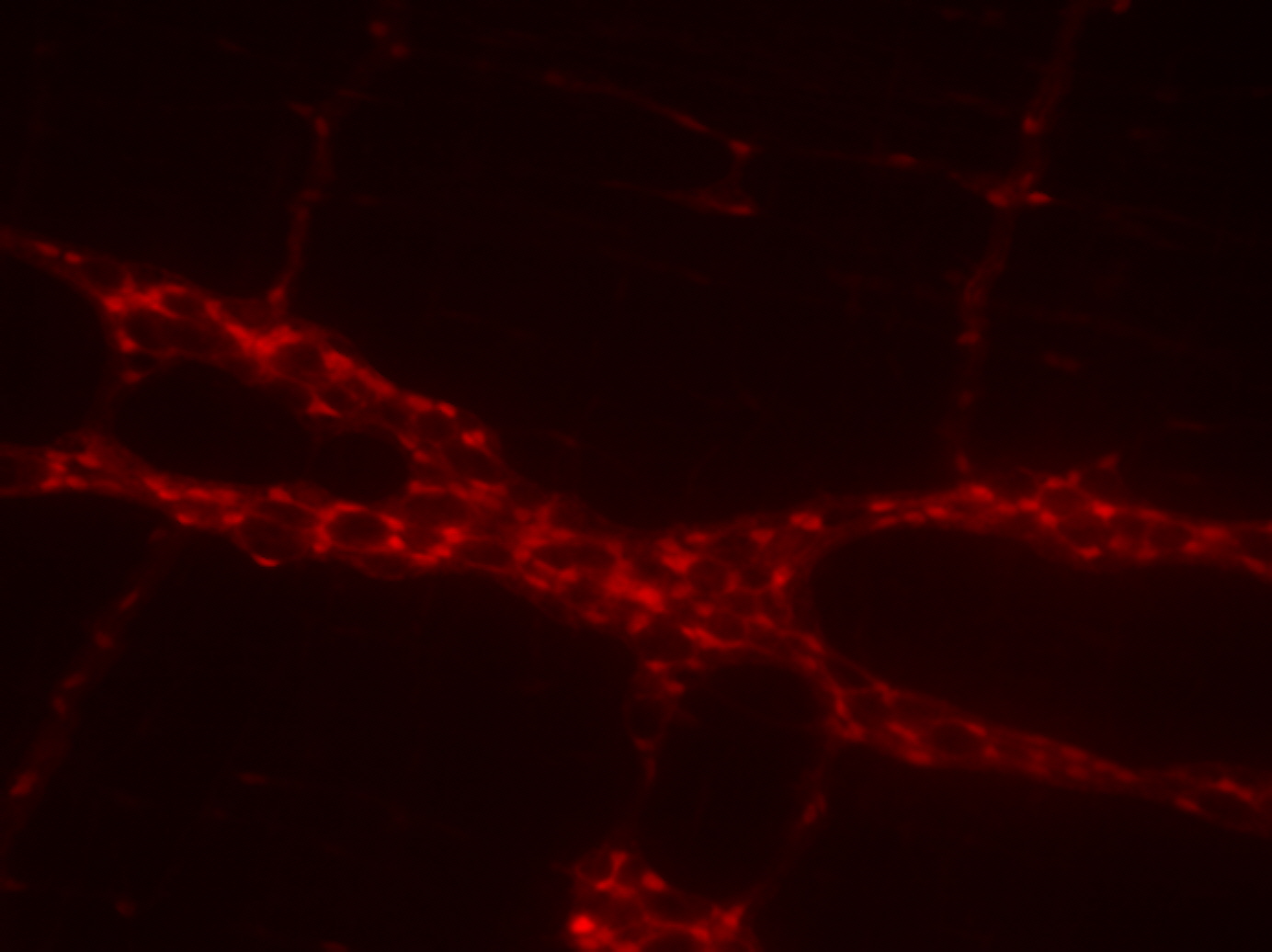}
        \caption{Original}
    \end{subfigure}
    \begin{subfigure}[h]{0.35\textwidth}
        \includegraphics[width=\textwidth, keepaspectratio]{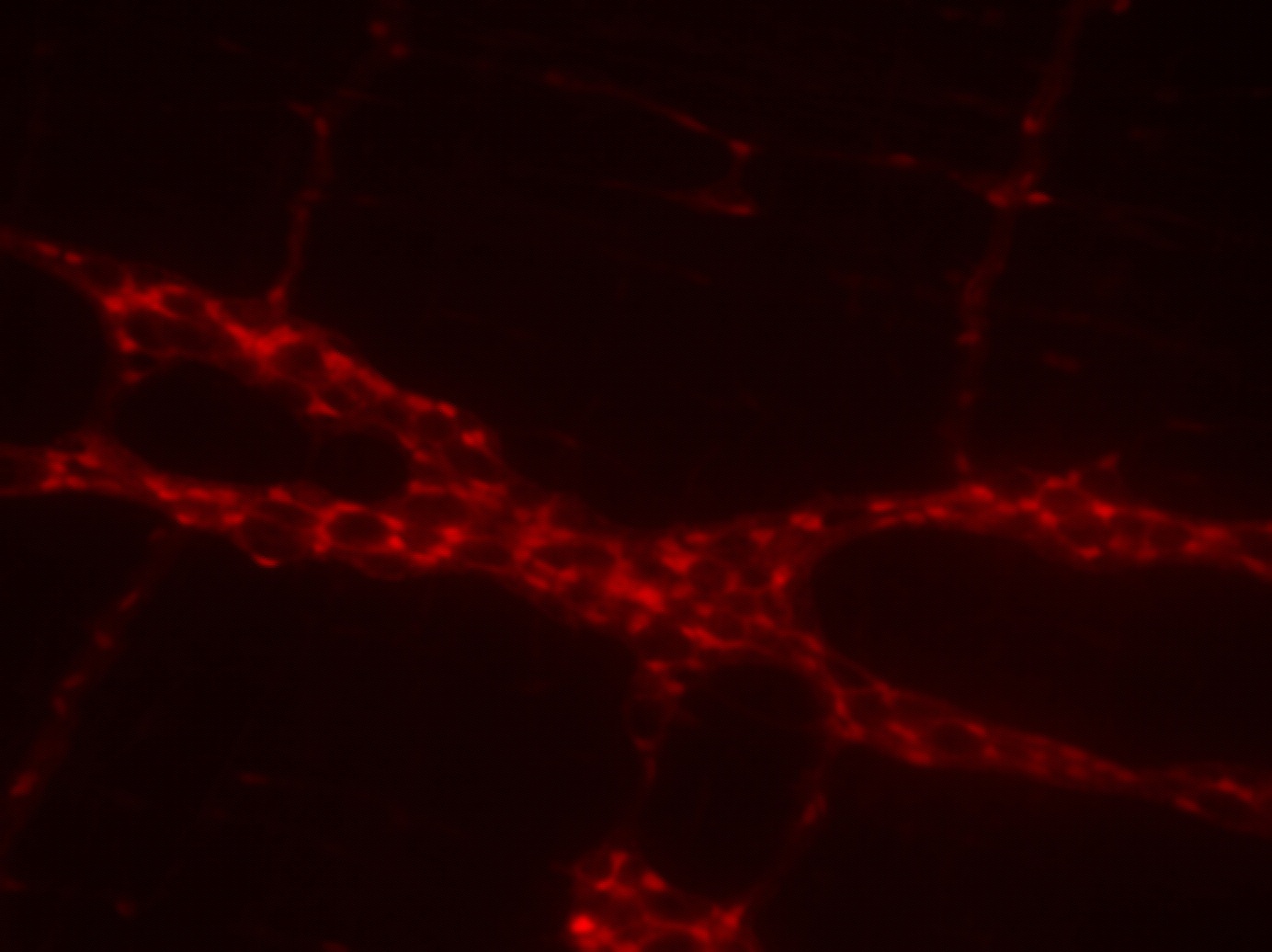}
        \caption{Laplacian}
    \end{subfigure}
    \begin{subfigure}[h]{0.35\textwidth}
        \includegraphics[width=\textwidth, keepaspectratio]{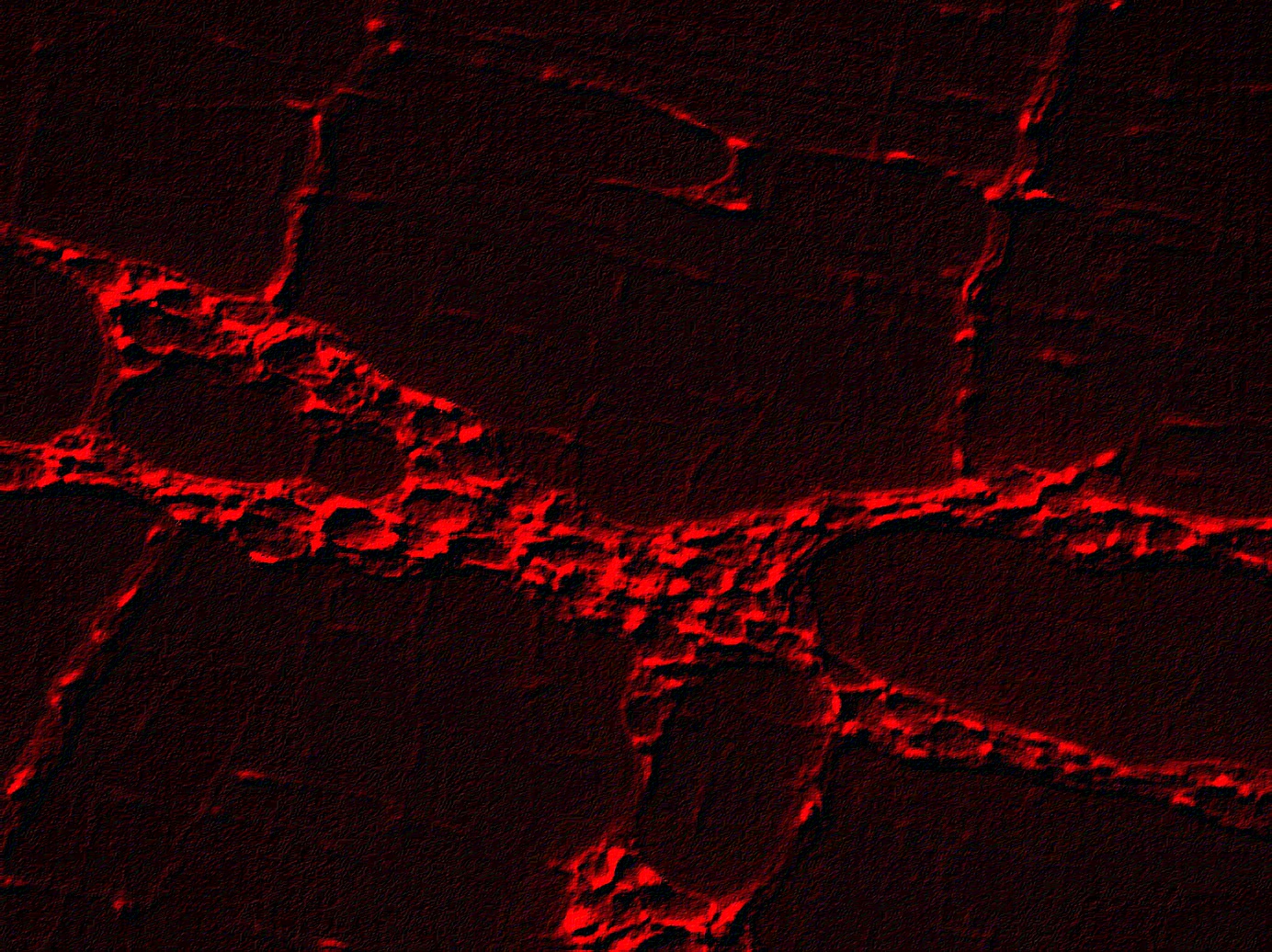}
        \caption{Scharr}
    \end{subfigure}
    \begin{subfigure}[h]{0.35\textwidth}
        \includegraphics[width=\textwidth, keepaspectratio]{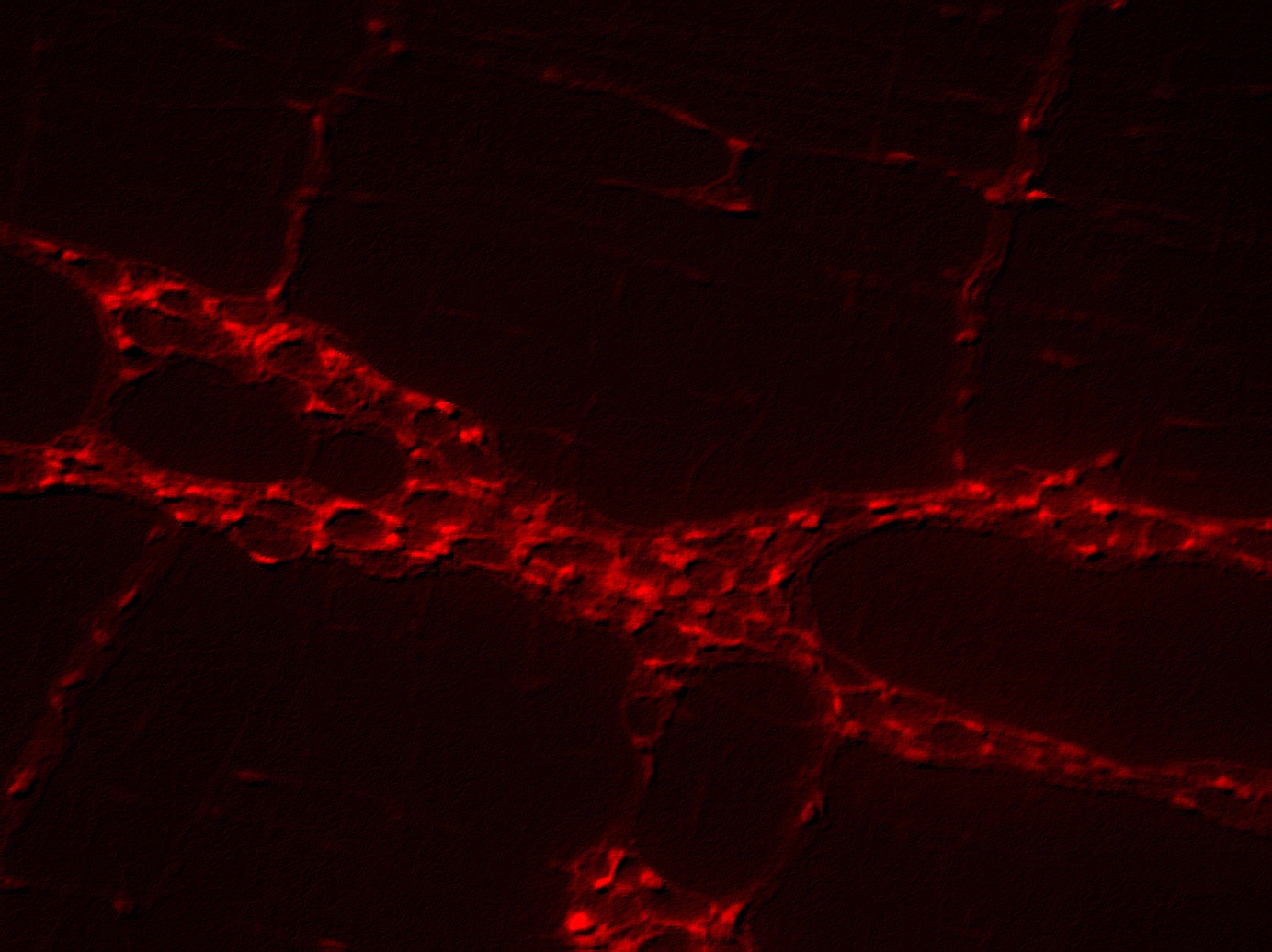}
        \caption{Sobel}
    \end{subfigure}
    \caption{Image sample from the C class, from the TW dataset. Sub-figure (a) presents the original data sample, while the remaining sub-figures show the resulting images by applying the edge highlighting methods using the following filters: Laplacias (b), Scharr (c) and Sobel (d).}
    \label{fig:gliaEdges}
\end{figure}

Considering that, the second goal aimed to reduce the existing blur in the image samples and increase the edges' definition. To achieve that, the data samples had their edges (or borders) highlighted, by detecting and adding them to the original image samples. This is made possible by edge detection methods, that generally apply a filter by using different kernels. In this work, three different filters were used: Laplacian, Sobel, and Scharr. Figure  \ref{fig:gliaEdges} presents a comparison between examples generated by using these filters.

    \subsubsection{Texture Descriptors}
            Different methods may be used to extract features from a given image. Regarding the visual attribute captured by these descriptors, we may observe that texture descriptors can potentially obtain good results in several different situations. And it stands out in different scenarios/applications, including in medical image analysis, biometric identification, etc.~\cite{maenpaa2003}.
    
The texture of a digital image is characterized by variations in the color intensity. By observing the differences between the pixels of the images, it is possible to provide a practical way to analyze an object's texture. Such analysis overlaps other ways to make an image's interpretation, e.g., by color.
    
 Among the different works described in the literature, carried out on diverse application domains, using texture based features extraction approaches, we can mention: music genre classification~\cite{costa2012music}, bird species classification~\cite{nanni_ICTAI_2016}, north atlantic right whale identification~\cite{freitas16}, identification of infants' cry motivation~\cite{felipe2019identification}, speech recognition~\cite{paulino2018brazilian}, acoustic scene classification \cite{felipe2017}, and COVID-19 identification using chest X-ray images~\cite{pereira2020covid}, among others.
    
 In this work, three handcrafted texture descriptor approaches were used: Local Binary Pattern (LBP), Robust Local Binary Pattern (RLBP), and Local Phase Quantization (LPQ).
    
The LBP was originally proposed by Ojala et al.~\cite{lbp} and uses a local neighborhood from every input pixel to generate a representative binary value. Two main parameters may be cited: $P$ and $R$. The $P$ parameter represents the number of neighbor pixels from a central pixel $c$, while $R$ represents the distance from it. The most common setup for LBP uses eight local neighbors ($P = 8$) two pixels distant from $c$ ($R = 2$)\cite{rlbp}. The representation of such configuration can be described as $LBP_{8,2}$ or $LBP(8,2)$. An extension of the original LBP defines the final feature vector as the normalized histogram that counts all uniform binary patterns (a binary pattern is considered uniform if it has no more than two transitions from 1 to 0 and vice-versa when evaluated as a circular list). This one has a total length of 59 features and presents better results when compared to the histogram of all individual binary patterns \cite{costa2013,rlbp,lbp}. More information about the LBP can be found in \cite{lbp}.
    
The RLBP was originally proposed by Zhao et al.~\cite{zhao2013completed} as a variation of the LBP texture descriptor. According to the authors, the RLBP is more accurate to capture the textural content from images that contain noise interference, when it results in non-monotonic gray level changes (even when such changes are not significant). The RLBP searches for a bit in the LBP pattern, that possibly suffered a variation inflicted by some kind of noise and then, review it. The original LBP's robustness is increased by this method, turning the binary patterns' uniformity concept a bit more flexible\cite{rlbp}. More details about the RLBP may be found in \cite{rlbp}.
    
The LPQ, was initially proposed by Ojansivu and Heikkil$\ddot{a}$~\cite{lpq}, being designed to be a texture descriptor more sensible to the image samples affected by blur. This descriptor has been presenting good performances, even in classification tasks, not targeted to images affected by this type of interference \cite{costa2013}. The LPQ uses periodic information from a bi-dimensional Discrete Fourier Transform (DFT), or specifically, a Short Term Fourier Transform (STFT). The STFT is computed in a rectangular neighborhood $N_x$ for each pixel in an image sample. The rectangular window size ($N_x$) is an important parameter to be varied, considering its direct impact in the generated features. After locally computing the texture of each pixel, the resulting codes are presented in a histogram, similarly as the LBP method. More detailed information about the LPQ texture descriptor can be obtained in \cite{lpq}.

    \subsection{Classifier algorithms}
         In the machine learning context, the supervised learning task builds a hypothesis capable of predicting an unobserved data sample label, based on the knowledge obtained from a known dataset, containing labeled data samples. In other words, given a dataset with $n$ examples of input/output pairs ($x_1$, $y_1$), ($x_2$, $y_2$), $...$, ($x_n$, $y_n$), in which $x$ and $y$ may represent any value (not necessarily numerical), it is built an hypothesis (function $h$) that approximates to a true function $f$ that generates each $y_i$ value starting from the $x_i$ value, i.e. $y=f(x)$. To evaluate this hypothesis, a set of test data samples is used.

This kind of task can be divided into two categories: regression problems and classification problems. If the output $y$ assumes a finite set of values, the task is categorized as a classification problem. Otherwise, if $y$ assumes a continuous numerical value the task is categorized as a regression problem \cite{norvig}.

 To the accomplishment of classification tasks, classification algorithms may be used. Different classification algorithms are described in the literature. In this work, four of them were approached: Support Vector Machines (SVM), Random Forests (RF), Na\"ive Bayes (NB) and k-Nearest Neighbors (k-NN).

 The SVM is a well-known method widely used by its efficiency to perform classification tasks. In its training step, a hyperplane is built, i.e., a decision limit with the shortest distance between the example points. In other words, it searches for a line (or surface) that segregates the patterns from different classes, being the margin defined as the distance from this one to the closest pattern. The support vectors are the transformed patterns that delimit such margin. The input is mapped to a higher dimensional space by a non-linear function, using what is known as Kernel Trick. Such action is performed based on the fact that some data may not be linearly separable in its original input space, but being easily separable when another dimension is added~\cite{Duda01,norvig}.
    
    The NB, can be categorized as a Bayesian Classification Algorithm. Such a category of algorithms, identify an object based on the posterior probability. Thereby, an object's class is assumed by the Bayes's theorem. Introduced originally by Thomas Bayes (1702-1761), the Bayes's Theorem (or the Bayes's Rule) is a simple equation, quiet often used by most modern artificial intelligence systems as a base for probabilistic inferences. It allows the calculation of a new unknown probability, by using three giving known conditional probabilities, that can usually be easily found. Based on this logic, the NB classifier assumes that a dataset's attributes are conditionally independent of each other. The prediction of an unseen data sample is then performed, based on the probabilities calculated from its attribute values, giving the labeled data samples, and targeting one specific class \cite{norvig}.
    
    The RF algorithm was originally proposed as a method of building classifiers based on decision trees, being as capable of increasing the accuracy in the training step as for samples not previously observed. Its operation can be shortly described as the build of multiple decision trees in a randomly selected subspace inside the features space. Then, generalizing the classification in different complementary approaches. By the end of the method's process, the tree-structured classifiers $h(x, \theta{i}), i \in \{1, .., k\}$, being $k$ equivalent to the total number of decision trees, cast individual votes for one of the possible classes of $x$. The final prediction is assumed to be the most popular class, i. e. the class with the greatest number of votes. More details about this algorithm may be found in \cite{tin95}.

The K-NN is a instance-based learning algorithm and has its operation based on the nearest neighbor rule. Considering a dataset with $n$ labeled samples $D^n = {x_1, ..., x_n}$, a certain $x'$ sample, such that $x' \in D^n$, can be described as being the closest point to a test sample $x$. By using the nearest neighbor rule, $x$ is classified with the same label/class as $x'$. This rule can be naturally extended to operate with a larger number of neighbors. In this way, $k$ nearest neighbors to $x$ are used to perform the decision when classifying such a test sample. In other words, each of the neighbors cast a vote for one of the possible classes, and at the end, $x$ is classified with the most popular class, i. e. the class with the largest number of votes. Considering this context, the traditional nearest neighbor rule is assumed as having $k=1$. It is noteworthy that to avoid draws, the value of $k$ always assumes an odd integer \cite{norvig,Duda01}.

\subsection{Non-Handcrafted Modeling}
    Traditionally, in the pattern recognition framework, features are extracted from the dataset and used as input in machine learning algorithms that are supposed to learn how to discriminate patterns from different classes. Thereby, a significant part of the effort put in works based on machine learning algorithms is dedicated to feature engineering. This is a time-consuming and challenging process that requires specialized knowledge.
    
That being said, we introduce the Feature Learning (FL), or Representation Learning (RL), concept. In this category, the techniques can automatically generate data representations, favoring the extraction of useful information during the build of classifiers and other predictive systems \cite{bengio2013}. Among the countless ways to perform FL, Deep Learning methods, such as Convolutional Neural Networks (CNN), can be highlighted. Keeping that in mind, this Section describes the non-handcrafted modeling, which aims to classify features extracted automatically by FL, being here referred to as non-handcrafted features.

    \subsubsection{Pre-Processing}
        Different approaches can be described to avoid overfitting, helping to ensure better and most accurate classification rates. One of them, Data Augmentation, artificially increases the total number of image samples by performing small changes in the original samples, using different operations~\cite{Srivastava14}. 

This approach is usually employed as a pre-processing technique when working with CNNs, taking into account the huge amount of data required to achieve satisfactory classification rates. Such a technique was applied in this work, considering the modest quantity of image samples existing in the used datasets.

\begin{figure}[htpb!]
    \centering
    \includegraphics[width=.85\textwidth, keepaspectratio]{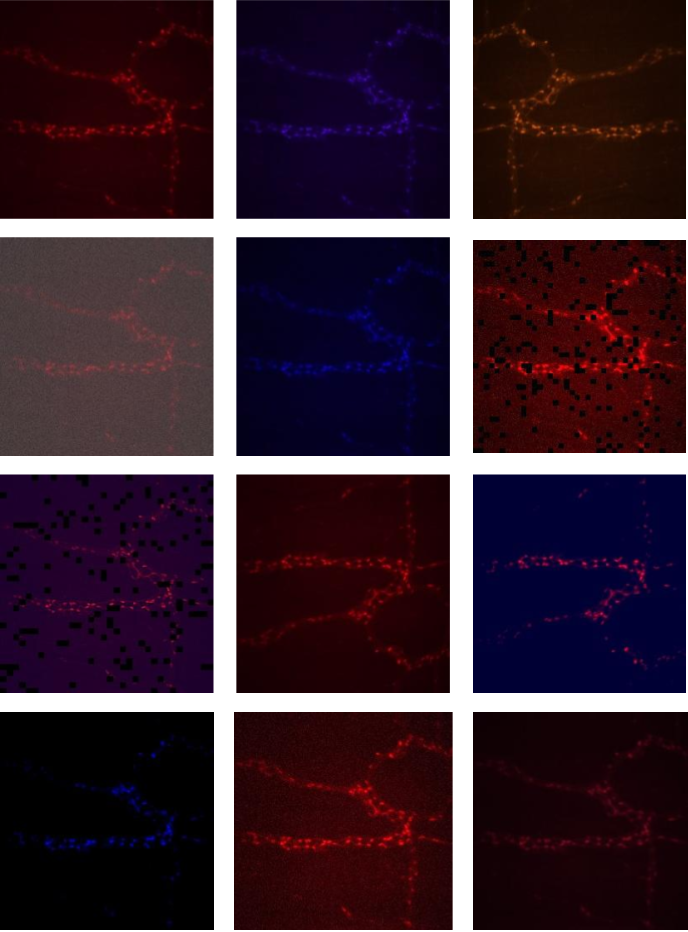}
    \caption{Examples of samples generated using Data Augmentation, by applying four randomly chosen techniques. Some of the techniques experimented are: adding Gaussian noise and/or blur; adding hue and saturation; flipping the image vertically and/or horizontally; changing the color space to BRG; Coarse Dropout; and others. The original image sample can be seen in the top left corner.}
    \label{fig:gc_da}
\end{figure}

To generate new image samples based on the ones available, randomly chosen image processing operations were applied to the original samples, such as: adding Gaussian noises, contrast normalization, adding blur noises, vertical/horizontal rotations, saturation variations, sharpen and others. Some examples of generated image samples can be seen in Figure \ref{fig:gc_da}.

    \subsubsection{Convolution Neural Networks (CNN)}
        CNN is an efficient algorithm, widely used in pattern recognition and image processing areas, due to its favorable characteristics, such as simple structure, less training parameters, and adaptability. Its shared weights structure makes it more similar to the neurons' connectivity pattern found in the human brain. Been heavily inspired in the human visual system structure, each neuron from a CNN does not globally visualize an entire image sample. But instead, only a portion of it (a local area) is visualized. This way, reducing the network model's complexity and the number of weights \cite{Liu2015,lecun89}.
 
A CNN can be considered a variation of the Multi-Layer Perceptron networks, been capable of applying filters in visual data, keeping the neighborhood relation between the image pixels through the network processing \cite{vargas16}.
 
    
Traditionally, the CNN's layers can be segregated in two sets: (1) the set of layers responsible for the image sample's feature extraction by FL (in this work, named non-handcrafted features), is usually composed of convolutional and pooling layers; and (2) the set of layers responsible for the classification, being composed of one or more fully connected (dense) layers~\cite{vargas16,matsu2018}. It is worth mentioning that the total number of layers may differ from one CNN network architecture to another.

    \subsection{Transfer Learning}
        Traditionally, machine learning algorithms predict the label of unknown data samples using models that were previously trained in a labeled dataset. With the evolution of such algorithms and the complexity growth of the tasks that employ them, it creates the need to heave an enormous amount of training data to achieve satisfactory results~\cite{Pan10}. 
        
The knowledge transferring concept induces the thought that a previous knowledge, acquired for a specific purpose, may be reused in another one. Having that in mind, the Transfer Learning ideal is presented. It aims to resolve new problems faster and with better solutions by using pre-obtained knowledge~\cite{Pan10}.

        
Transfer Learning extracts knowledge from a source task and uses it in a target task. When using CNNs, this method makes possible the use of pre-trained models (trained in different datasets/purposes) in new classification tasks. Thus, demanding a smaller number of data samples in CNN's training. Once the FL layers are kept unchanged, only requiring the classifications layers to be trained. It is worth mentioning that the quantity and dimensions of such layers may differ from those used originally in the base model, aiming to adapt the network to the target task. Also, it can be replaced by traditional classification algorithms, e.g., Support Vector Machines (SVM).
         
Finally, it is essential to emphasize that the use of transfer learning in this work can be seen as a timely strategy to deal with issues regarding the dataset's modest size. In our specific application, it is not easy to enlarge the dataset since it needs new animal resources and the appliance of a long-term method.

\subsection{Classifier Combination}
    Many classification algorithms used to generate probability estimates in their outputs. Such values reference the prediction scores for each class present in the problem, obtained from the evaluation of test samples. It is possible to perform different operations over these probability scores aiming to combine them. This way, new predictions scores are obtained based on the previously reached values, generating a final result.
    
Three classifiers combination rules, originally proposed by Kittler et al.~\cite{kittler1998}, were used in this work for such finality. In these equations, $x$ refers to the pattern to be classified, $n$ is the number of classifiers involved in the combination, $c$ is the number of classes, $y_i$ is the output label of the $i_{th}$ classifier in a problem with the following possibilities of classes labels $\Omega = \omega_1, \omega_2, ..., \omega_c$, and $P(\omega_k | y_i(x))$ is the probability of the sample $x$ to belong to the class $\omega_k$ according to the $i_{th}$ classifier.
    
\begin{itemize}
	\item Max Rule: for each existing class from the dataset, the maximum value between the prediction scores from different classifiers is chosen. Later on, the final result is given by the class with the biggest score. This combination rule can be represented by Equation \ref{regraMax}.
	       
	\begin{equation}\label{regraMax}
        max(x)=\argmax_{k=1}^c max_{i=1}^n P(\omega_k | y_i(x))
    \end{equation}	       
	       
	\item  Sum Rule: considering all generated classifiers, the calculated predictions are summed for each class. Then, the class with the maximum score is chosen as the final result. This combination rule can be represented by Equation \ref{regraSoma}.
	       
	\begin{equation}\label{regraSoma}
	    sum(x) = \argmax_{k=1}^c \sum_{i=1}^n P(\omega_k | y_i(x)) 
	\end{equation}
	       
	\item  Product Rule: this rule works similarly as the sum rule. But, instead of performing a sum operation, the values are multiplied. This combination rule can be represented by Equation \ref{regraProd}.
	       
	\begin{equation}\label{regraProd}
        prod(x) = \argmax_{k=1}^c \prod_{i=1}^n P(\omega_k | y_i(x))
	\end{equation}
\end{itemize}

It is important to remember that by combining classifiers, we have an opportunity to merge results obtained using handcrafted and non-handcrafted features once we have probabilities predictions as output from classifiers on both modes. The combination of handcrafted and non-handcrafted approaches aiming to get a single final decision has somehow already proven to be effective in other works~\cite{costa2017evaluation,nanni17}.

\section{The Datasets}
    \label{sec:database}

\begin{figure}[htpb!]
    \centering 
    \begin{subfigure}[b]{0.3\textwidth}
        \centering 
         \includegraphics[width=\textwidth]{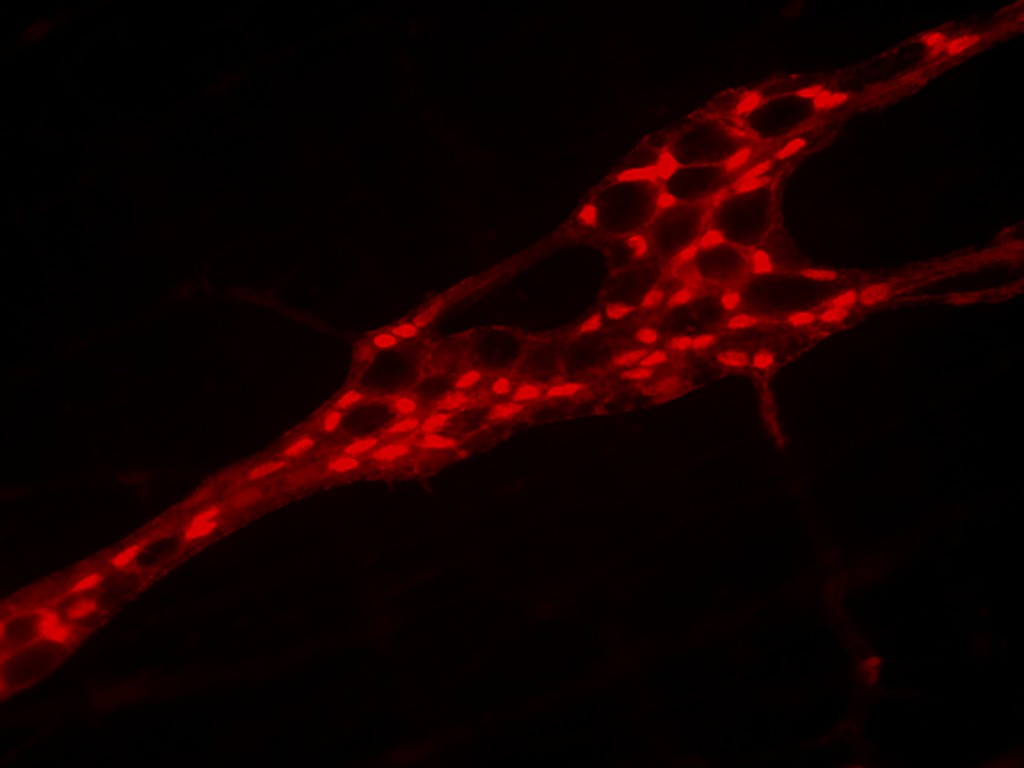}
         \caption{AIA dataset  \& C class}
         \label{fig:aia_c}
    \end{subfigure}
    \begin{subfigure}[b]{0.3\textwidth}
        \centering 
         \includegraphics[width=\textwidth]{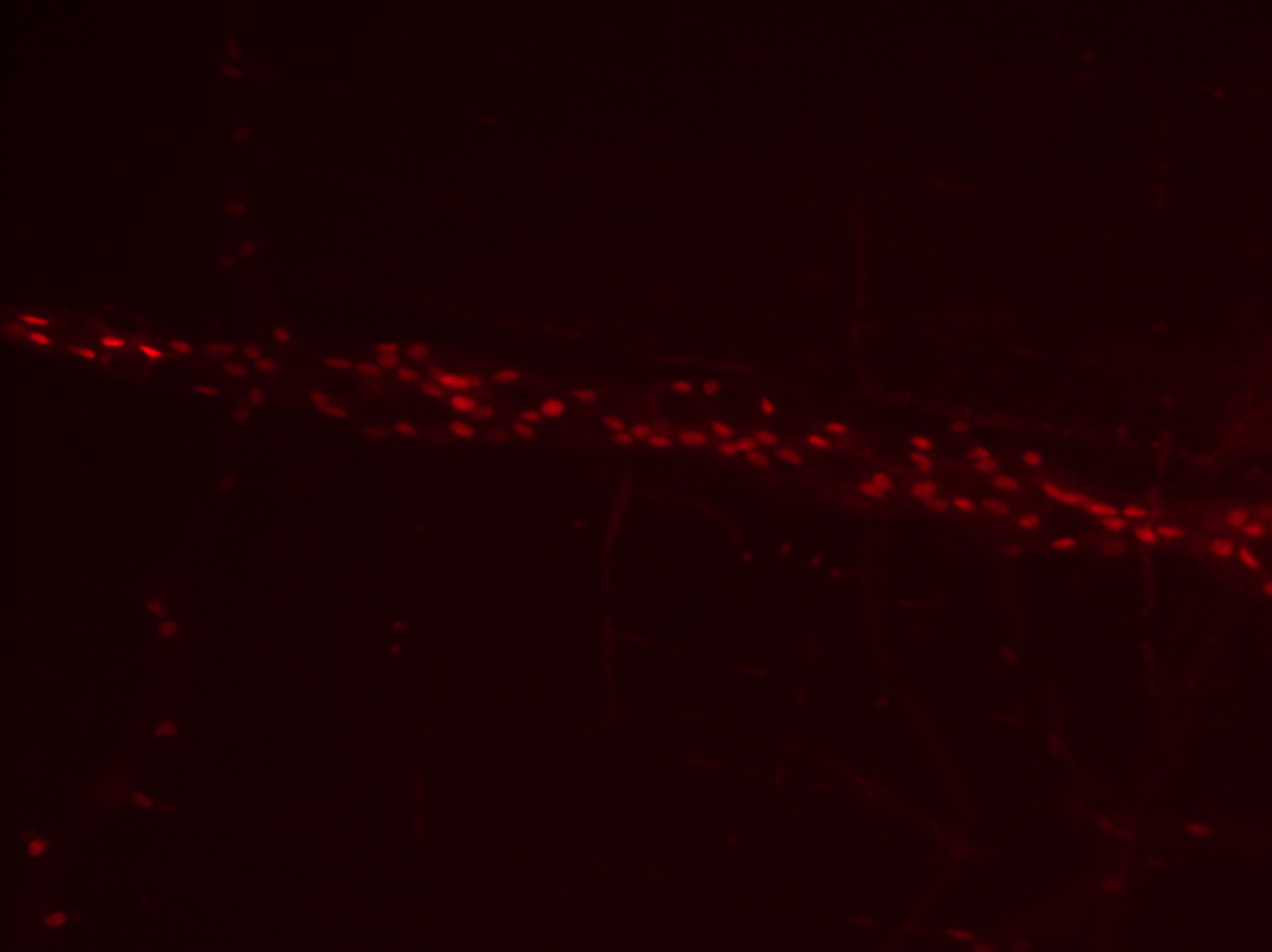}
         \caption{D dataset \& C class}
         \label{fig:d_c}
    \end{subfigure}
    \begin{subfigure}[b]{0.3\textwidth}
        \centering 
         \includegraphics[width=\textwidth]{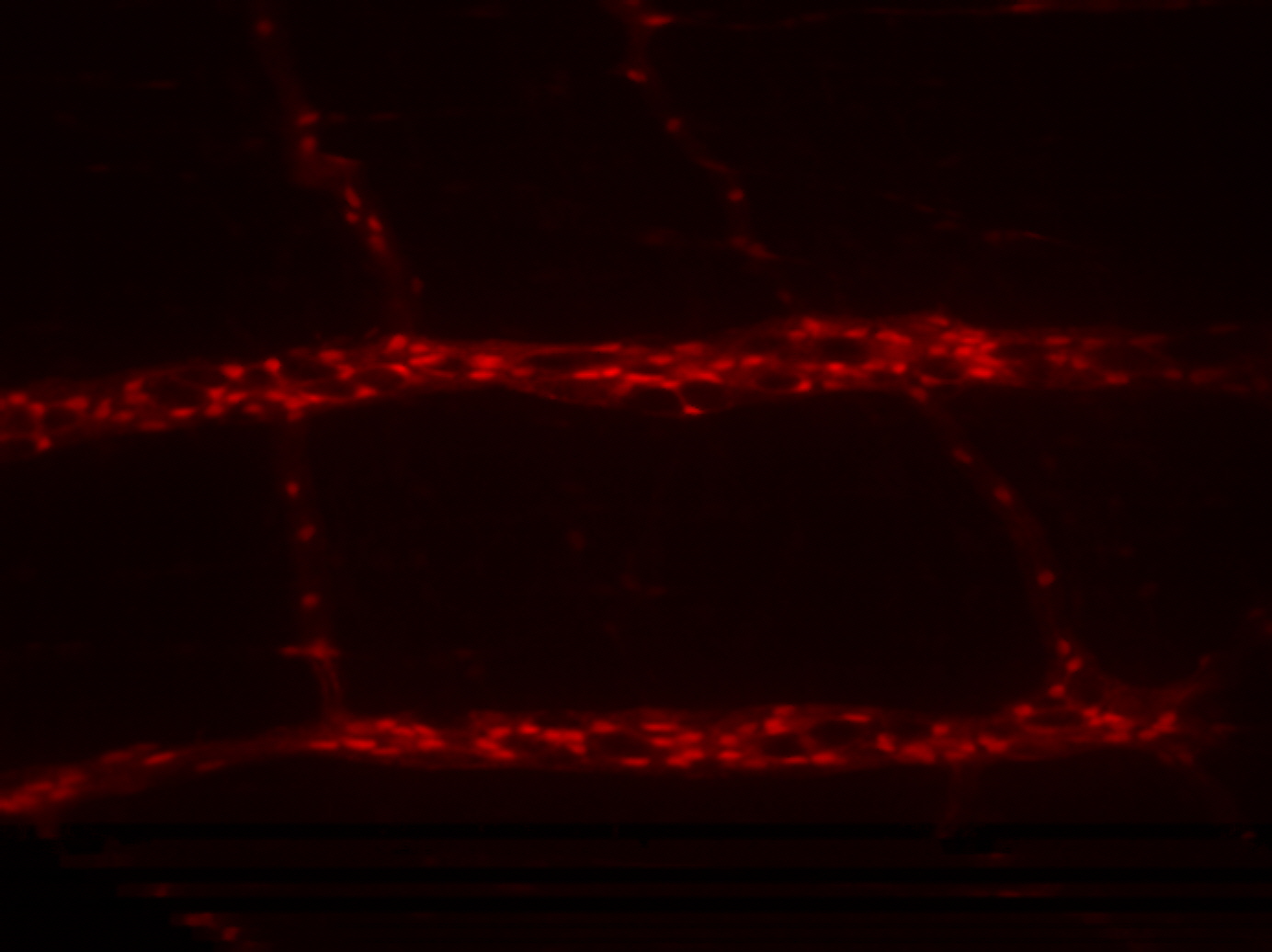}
         \caption{TW dataset \& C class}
         \label{fig:tw_c}
    \end{subfigure}
    \begin{subfigure}[b]{0.3\textwidth}
        \centering 
         \includegraphics[width=\textwidth]{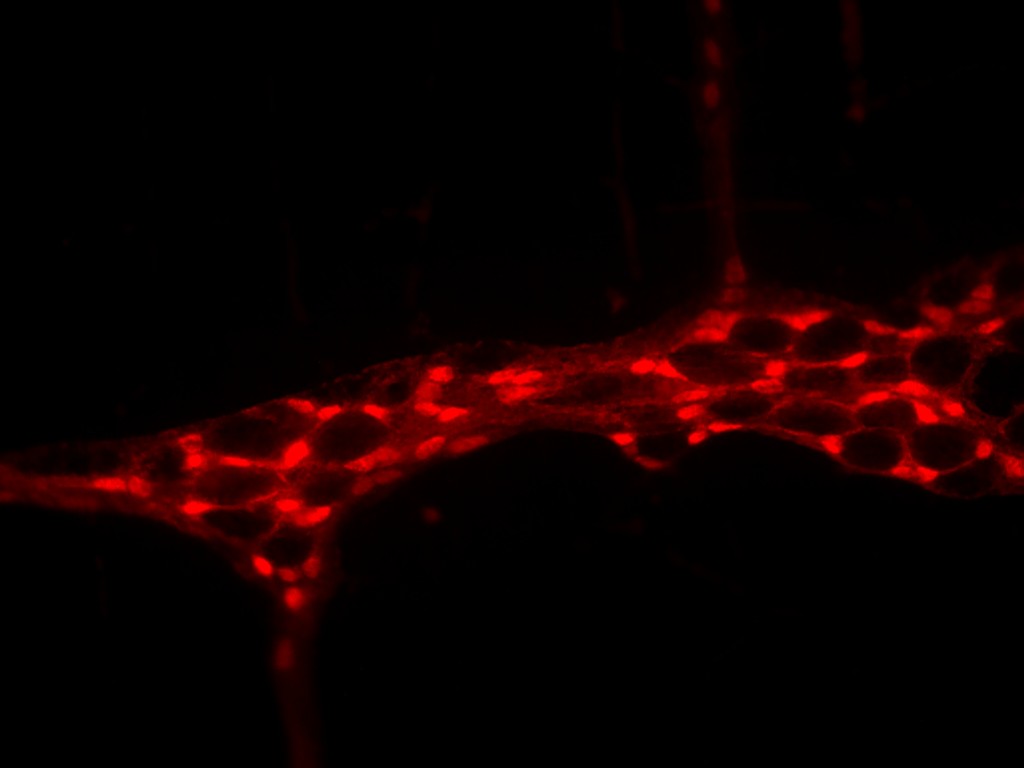}
         \caption{AIA dataset \& S class}
         \label{fig:aia_s}
    \end{subfigure}
    \begin{subfigure}[b]{0.3\textwidth}
        \centering 
         \includegraphics[width=\textwidth]{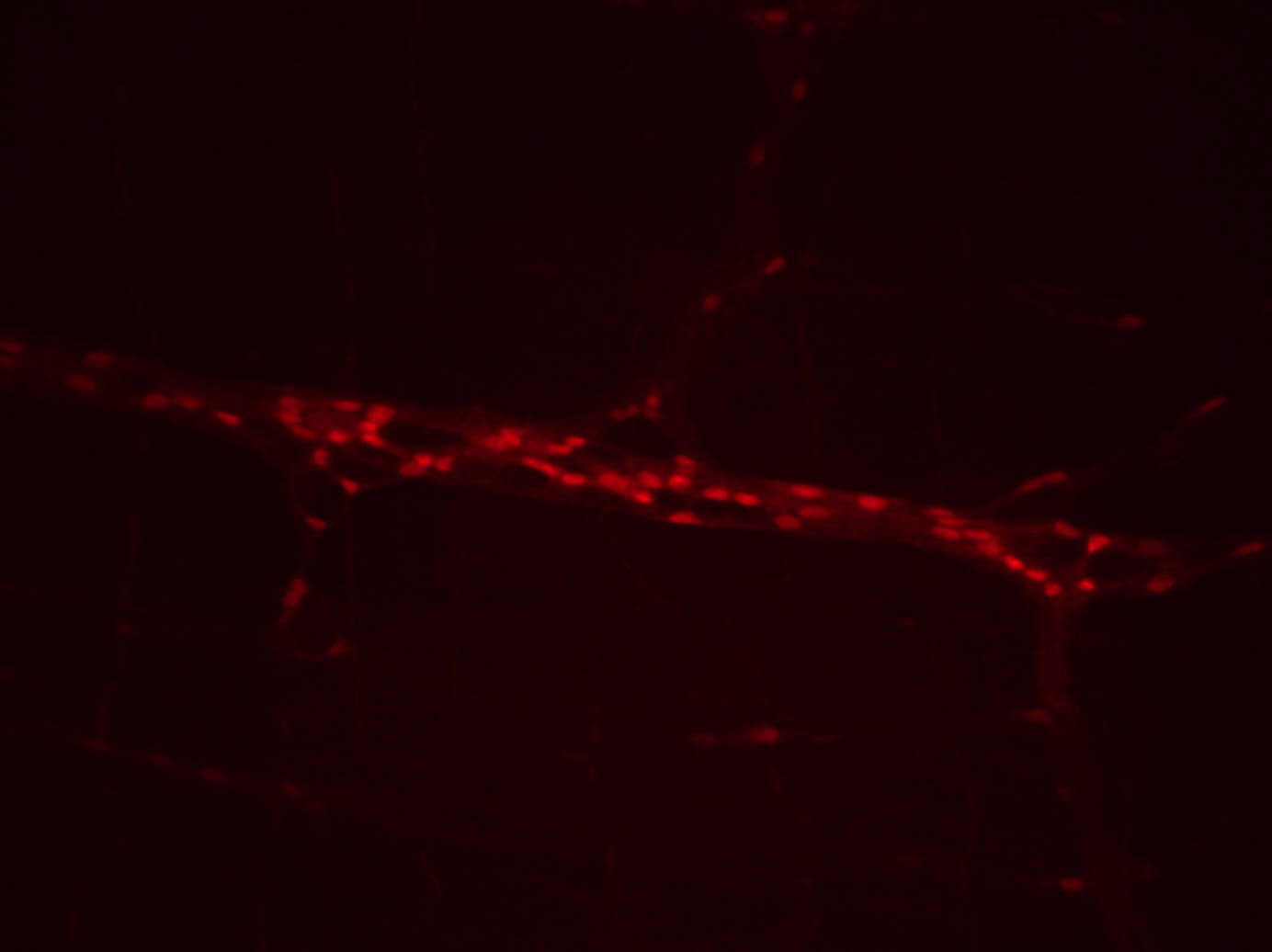}
         \caption{D dataset \& S class}
         \label{fig:d_s}
    \end{subfigure}
    \begin{subfigure}[b]{0.3\textwidth}
        \centering 
         \includegraphics[width=\textwidth]{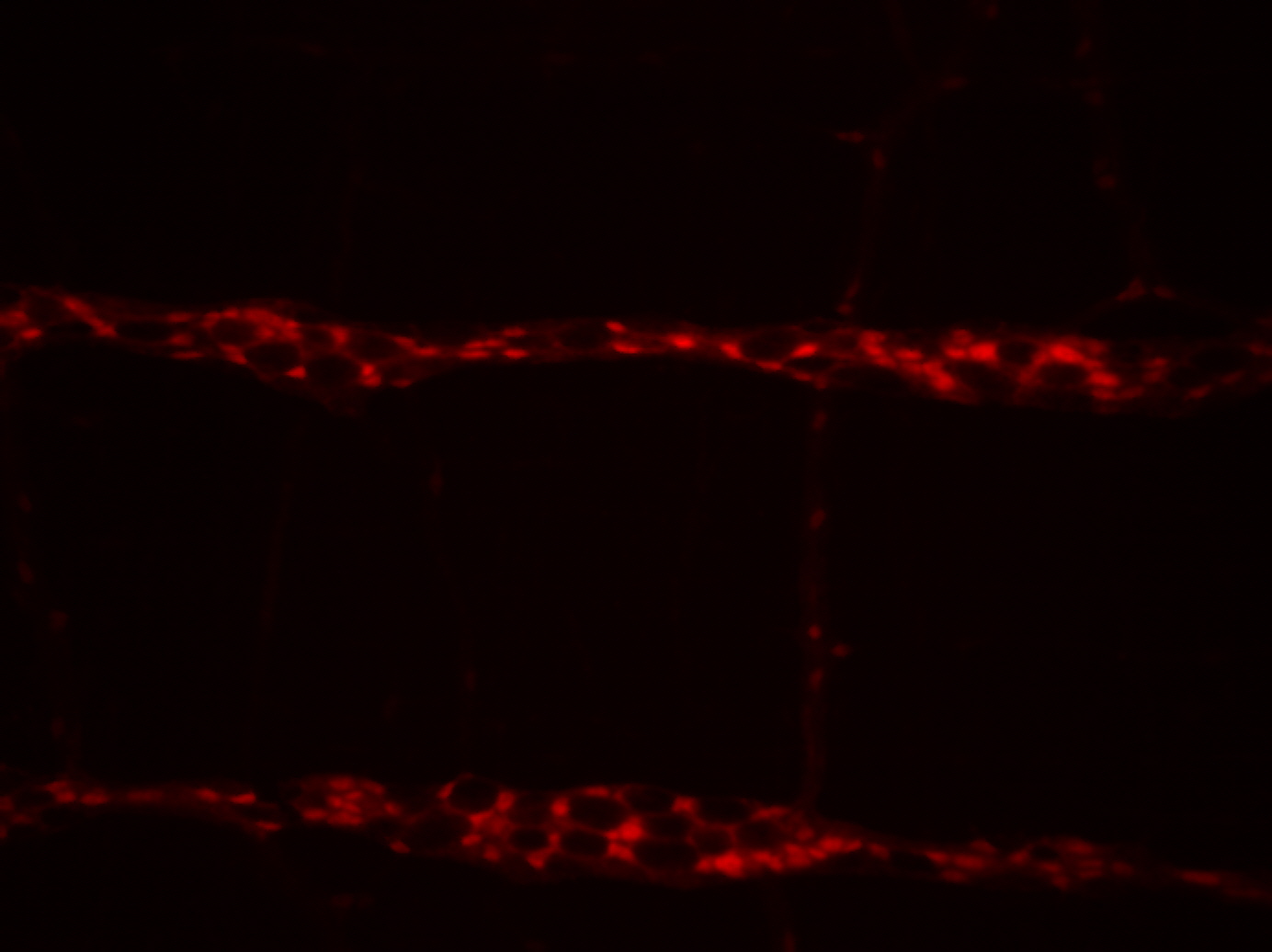}
         \caption{TW dataset \& S class}
         \label{fig:tw_s}
    \end{subfigure}
    \caption{Image samples from the datasets and both classes. The red coloration is resulted from the immunostaining of the S100 protein. The EGC can be visualized as the most bright points in the samples.}
    \label{fig:image_sample}
\end{figure}

This work presents three novel datasets created by researchers of the Enteric Neural Plasticity Laboratory of the State University of Maring\'a. Such datasets are composed of image samples obtained from the ENS of rats, in which EGC can be visualized through the immunostaining of the S100 protein. Figure \ref{fig:image_sample} shows some image samples taken from the datasets. Each dataset represents an investigated disease, being them: arthritis rheumatoid (AIA), cancer (TW) and diabetes mellitus (D). 

Each dataset is represented by the diseases' name abbreviations (AIA, TW, and D) in this work. It is worth mentioning that in this scenario, the datasets can be also called ``disease groups''. The datasets are composed of two classes, one containing image samples extracted from sick animals (S) and other from control/healthy (C) ones. The exact quantity of image samples per dataset and per class, and the image samples' dimensions, can be seen in Table \ref{tab:classes_details}.

The datasets were created taking into account the ethical principles under the terms set out in the Brazilian federal law\footnote{Law 11,794 (October 2008) and the decree 66,689 (July 2009).}, established by the Brazilian Society of Science on Laboratory Animals (SBCAL). All the proceedings were submitted and approved by the Standing Committee on Ethics in Animals Experimentation of the State University of Maring\'a\footnote{Protocol numbers 062/2012 (TW), 4462180216 (AIA) and 073/2014 (D).}. After the experimental time, the animals were frozen and sent to incineration.

\begin{table}[ht]
		\centering
		\caption{Diseases evaluated in this work, according to the categories' distribution in the experimental groups.}
		\resizebox{\textwidth}{!}{
		\scalefont{0.7}
		    \begin{tabular}{c c c c c}
		        \hline
			    \multirow{2}{*}{Disease} & \multirow{2}{*}{Abbreviation} & \multirow{2}{*}{Image Dimension} & \multicolumn{2}{c}{Number of Samples}\\\cline{4-5}
				    & & & Sick (S) & Control (C) \\
				    \hline
				    Arthritis Rheumatoid & AIA & 1024 x 768 & 210 & 208 \\
				    Cancer (Walker\textquotesingle s tumor-256) & TW & 1384 x 1036 & 192 & 224 \\
				    Diabetes Mellitus & D & 1384 x 1036 & 290 & 224\\
			    	\hline
			\end{tabular}}
		\label{tab:classes_details}
\end{table}

The jejunum, i.e., the second part of the small intestine, of male adult Wistar rats (Rattus norvegicus), albius variety (D and TW) and holtzmann rats (AIA) were used in this study. The experimental models were developed according to the works of Frez et al. (2017) \cite{frez2017} (D), Vicentini et al. (2017) \cite{vincentini17} (TW) and Souza et al. (2011) \cite{souza11} (AIA). Variations in the animals' ages were encountered as well, in every disease investigated here. The animals' euthanasia was performed with tiopental (150 mg/kg)\footnote{Abbot Laboratory, Chicago, IL, USA} intraperitoneally. Then, the jejunum was collected and processed to evidence the EGC.

The jejunum fixation and S100 immunostaining, were performed according to the protocol proposed by Pereira et al. (2011) \cite{pereira11}. The images of the EGC were obtained using an optic microscope\footnote{Olympus BX 41, Tokyo, Japan} with immunofluorescence filters and high-resolution camera\footnote{Moticam\textregistered\space 2500 5.0 Mega Pixel - Motic China Group Co., Shanghai, China} attached to a computer. Photomicrographs were recorded using Motic Images Plus 2.0ML software (Motic China Group Co.). The image samples were then obtained using 20x, from randomly chosen places of the animals\textquotesingle jejunum.

It is worth mentioning that the image samples' acquiring is variable according to the target dataset. It can be justified mainly by the possibility of distinct tissue fixations and specific immunostaining responses. More detailed information about these processes and the datasets, such as the induction of the disease, may be found in \cite{frez2017} (D), \cite{vincentini17} (TW), and \cite{souza11} (AIA).

The datasets used in this work were made freely available for research purposes\footnote{\url{https://github.com/gustavozf/EGC_Z_dataset}}, such a way that other researches can take benefit from it, and properly compare the results obtained using different techniques with those obtained here.

\section{Experimental Analysis}
    \label{sec:results}
In this section we describe the results obtained after performing the experimental protocol designed in this work. Subsection \ref{subsec:ei} describes the exploratory investigation taken that leaded to the achievement of this work's best classification rates. Subsection \ref{subsec:pc} presents the parameters and implementations used in the algorithms employed in this work. Finally, Subsection \ref{subsec:res} presents and briefly discusses this work's results.

\subsection{Exploratory Investigation}\label{subsec:ei}
    \label{sec:exp_inv}
The experiments were performed, fallowing a result-oriented exploratory research approach. Such experiments may be divided into three categories: (1) the ones that evaluated the performance of handcrafted features; (2) the ones that evaluated the performance of non-handcrafted features; and (3) the ones that combined the resulting classifiers from the previous categories.

\subsubsection{Handcrafted modelling}
The set of experiments using handcrafted features was designed to evaluate different approaches, such as assessing different texture descriptors, enhancing the image samples' visual features, and testing different classification algorithms. 

The first experiments aimed to evaluate the performance of features extracted based on the image samples' texture. To accomplish such a goal, we used LBP, RLBP, and LPQ texture descriptors. These were applied to the image samples, and the resulting features were used as input in the SVM classification algorithm. It is worth mentioning that the assumed values of the following parameters: neighborhood size, distance from the central pixel (LBP and RLBP), and window size (LPQ), were chosen according to the best performances obtained by Costa \cite{costa2013}. 

From these experiments' results, it was possible to notice a tendency to find more significant results by using the LPQ texture descriptor and inducing that it would be the most appropriate for the here approached problem. This may be justified by the fact that the LPQ was created to better extract features from image samples affected by blur, a kind of signal noise present in the image samples of this work's dataset. It is worth mentioning that the AIA's classification performance reached its best values by using the RLBP texture descriptor. That can be justified because the image samples from this disease group are less affected by blur and the ones more affected by the inconvenience of different noises generated by the immunohistochemistry background.
        
Having that in mind, new experiments were performed to reach better classification rates. To achieve this goal, new features were extracted from the image samples, using the LPQ texture descriptor and making variations the window size value, assuming the values 3, 5, 9, 11, and 13. From these results, it is possible to observe better classification rates than the previously performed experiments. Showing that there is still a space for improvement of the LPQ performance on this task if the parameters are properly adjusted.

From this point, we aimed to evaluate the performance of experiments carried out by applying pre-processing techniques. Firstly, the image samples had their colors omitted by converting them to grayscale to achieve this goal.  Since these kinds of operations do not affect the existing texture in the image samples, the texture descriptors, and their parameters, used in these experiments were the ones that reached the best classification rates until this moment. The resulting F-measure values found from these experiments showed an improvement on the AIA and D datasets.

In a second moment, the image samples were pseudo-colored. Different color maps were evaluated to perform the pseudo-coloring, such as HSV, Autumn, Jet, Rainbow, and others. The resulting images were observed, aiming to search for the best choice that created greater visibility of the EGC. From the ones tested, it was possible to note that the HSV colormap best suited the searched goal.

By analyzing the results obtained from this experiment, we observe an opposite behavior compared to the results that converted the image samples to the grayscale. For the AIA and TW datasets, the found F-Measure values were inferior to the ones found in past experiments. While for the D disease group, a slight improvement could be observed compared to the best one found until this point.

Then, considering the existing blur in the image samples, new experiments were performed aiming to reduce this kind of noise and increasing the definition of the edges. The data samples had their edges (or borders) highlighted to achieve this goal by detecting (with the Laplacian, Sobel, and Scharr filters) and adding them to the original image samples. For this approach, the LBP, RLBP, and LPQ texture descriptors were used to extract handcrafted features.

The choice for re-testing the LBP and RLBP texture descriptors in these experiments, even knowing the superior performance of the LPQ in this problem, is justified by the direct impact that the edge highlighting method generates in the image's texture. This creates the need to reevaluate the performance of these texture descriptors, giving the new environment to be explored. Besides, in this case, the test is performed with images supposedly free from undesirable effects of blurring (a characteristic that the LPQ excels at). The operations were applied in the entire dataset. From the new image samples generated, features were extracted using the texture descriptors previously mentioned and classified by the SVM classification algorithm.

By the end of these experiments, it is possible to conclude that although the image samples generated by edge highlighting methods may be visually better when used as input in classification systems, the values obtained from the classification rates were inferior to those obtained without such process. One may conclude that such an approach would be inefficient for the problem investigated here.

The experiments performed until this point, aimed to identify the best possible way to represent the data samples of the approached problem, used as input through handcrafted features. In such experiments, the SVM classification algorithm was always used. Therefore, we carried out new experiments to evaluate different classification algorithms' performance, using best-extracted features found, i.e., features that led to the best classification rates until now, as input to the algorithms. To achieve this goal, the k-NN, RF, and NB classification algorithms were tested. However, the results found during these experiments' execution were still inferior to the ones found using the SVM classification algorithm. 

\subsubsection{Non-handcrafted modelling}
From this point, the methods that use non-handcrafted features, i.e. features extracted automatically by FL, are evaluated. Therefore, experiments were performed using different CNN architectures and transfer learning.

The CNN tests were performed using the LeNet5, AlexNet and MaxNet CNN architectures. From these experiments, it was possible to conclude that, although the number of samples in the dataset was artificially increased, the experimental models' training was not able to succeed in the task of abstracting the problem in question.

New experiments were them employed, aiming to investigate the efficiency of applying the Transfer Learning technique in the problem investigated here. At the end of the proposed experiments, it is possible to notice good performance rates obtained by applying the Transfer Learning technique. These rates reached equivalent or slightly greater values when compared to the ones resulted by classifying handcrafted features. Showing that such a technique would be more appropriate for the classification of two out of three datasets addressed here, AIA and D. Justified by the fact that the strategy accomplished here does not demand a huge amount of data samples, like traditional CNN classification methods do, and extracts features using pre-trained FL layers proposed to solve problems from different domains.

\subsubsection{Classifiers Combination}
Previously performed experiments used handcrafted and non-handcrafted features as input to create classification models. By such experiments, it was possible to observe the performance of different feature extraction and classification techniques to ascertain the generated classifications rates' behaviors on the investigated problem. With this in mind, the last experiments performed aimed to combine the most promising classifiers presented until this point for each scenario, i.e., handcrafted and non-handcrafted.

For each dataset, six classifiers were chosen, three of them obtained from handcrafted features, and the other three from non-handcrafted features. Thus, all possible sets of configurations were generated, which is equivalent to 57 different possibilities ($2^6 - 7 = 57$, ignoring the empty set and the ones with length equal to one) for each disease group. These were combined using the Sum, Product, and Max rules. At the end of these experiments, one may notice an improvement in the classification rates for every disease group, which indicates a complementarity between the classifiers (handcrafted and non-handcrafted) used.

\subsection{Parameters and Configurations}\label{subsec:pc}
    The image samples were pseudo-colored using the color-maps made available by the OpenCV library\footnote{\url{www.opencv.org}}. The edge enhancement and the image samples' conversion to the grayscale were performed using such a library.

The SVM implementation used here belongs to the libSVM library~\cite{libsvm}. We have used the Radial Basis Function (RBF) kernel, and a grid search procedure optimized the parameters $C$ and $\gamma$. The remaining classification algorithms implementation used here are those available in the Scikit-learn \cite{scikit-learn} library. The k-NN algorithm used five nearest neighbors ($k=5$) to perform the classification. The distance was calculated by the Minkowski Distance with $p=2$ and the voting system had uniform weights, i.e., the neighbors' votes had the same weight in the final prediction. The RF algorithm used ten trees, and the Gaussian version of the NB algorithm was used. Such parameters were chosen considering the default values suggested by the \textit{scikit-learn} library.

The CNNs used were implemented by the use of the Keras \cite{2015keras} library. The created models were compiled using the Adam Optimizer \cite{adamOpt}, with the $\beta_1$, $\beta_2$ and decay parameters equivalent to 0.9, 0.999 and 0.0005 respectively. 

The CNN tests were performed using the LeNet5, AlexNet, and MaxNet CNN architectures, with batch size equal to 128 and learning rate varying between $10^{-3}$ and $10^{-5}$. The number of epochs was set to 1024. It is worth mentioning that the model's performance through the epochs was monitored by the use of early stopping, with the patience of 64 epochs. After the training step, the best model found in the iterations was saved (checkpoint) and used in the test step.

To reduce the possibility of overfitting in the CNNs that did not approached Transfer Learning, two different techniques were applied: (1) artificially increasing the number of image samples through data augmentation, generating 32 new image samples from each of the existing ones; and (2) using dropout layers in the CNN architectures.

The Transfer Learning experiments used the FL layers of the VGG16, InceptionV3, and InveptionResNetV2 architectures to extract features of the EGC images. It is worth mentioning that these layers' weights were obtained by performing the training in the ImageNet dataset. The implementations os such networks are all made available by the Keras library. Unlike common approaches, instead of such features being classified by a new set of fully-connected (dense) classification layers, the SVM classification algorithm was used for such a task. This is justified by the results previously described in this study.

Additionally, considering that the FL layers of the CNN architectures that approached the Transfer Learning technique, generate a great number of features, the Chi-Square ($\chi^2$) statistical test was employed as a Feature Selection technique, aiming to reduce the number of features to a smaller value $n$. Being the tested values for $n$: 256, 512, 1024, 2048, and 4096 features.
    
\subsection{Results}\label{subsec:res}
This section presents the classification rates found when executing the experiments accordingly to the exploratory investigation presented in Section \ref{sec:exp_inv}.

\subsubsection{Handcrafted modelling}
Table \ref{tab:desTexInic} presents the results found in the experiments that aimed to evaluate the performance of features extracted by texture descriptors. It can be observed that for the AIA disease, the best results\footnote{In this work, the best results will always be defined by the best F-measure performance, once this metric is given by the harmonic mean between recall and precision.} were found when the RLBP texture descriptor was used, with a F-Measure of 0.8062. For the remaining diseases, such results were found by using the LPQ texture descriptor, reaching F-Measures of 0.9447 and 0.8637 for TW and D diseases, respectively.

\begin{table} [htpb!]
    \caption {F-Measure values found from the experiments using different texture descriptors.}
    \begin{center}
        \begin{tabular}{c c c c} \hline
            \multirow{2}{*}{Dataset} & \multicolumn{3}{c}{Texture Descriptor}\\\cline{2-4}
                & LBP$_{8,2}$ & RLBP$_{8,2}$ & LPQ$_7$\\\hline
            AIA & 0.7895  & \textbf{0.8062} & 0.7535\\
            TW  & 0.9279  & 0.9231 & \textbf{0.9447}\\
            D   & 0.8520  & 0.8327 & \textbf{0.8637}\\
             \hline
        \end{tabular}
        \label{tab:desTexInic}
    \end{center}
\end{table}

Table \ref{tab:lpqWinSize} reports the experiments' classification rates that extracted features using the LPQ texture descriptor and used variations in the window size value. By analyzing these experiments, it was possible to observe a better efficiency of features extracted by the LPQ texture descriptor. Table \ref{tab:bestOfHcf} compares the best results.
        
\begin{table}[htpb!] 
    \caption {F-measure values found from experiments that explored different values for the LPQ's window size.}
    \begin{center}
        \begin{tabular}{c c c c c c} \hline
            \multirow{2}{*}{Dataset} & \multicolumn{5}{c}{Window Size}\\\cline{2-6}
            & 3 & 5 & 9 & 11 & 13 \\\hline
            AIA & 0.7751 & 0.7679 & 0.7943 & \textbf{0.8110} & 0.8014\\
            TW  & 0.9447 & 0.9591 & 0.9447 & 0.9567 & \textbf{0.9712}\\
            D   & 0.7838 & 0.8189 & \textbf{0.8773} & 0.8690 & 0.8636\\
            \hline
        \end{tabular}
    \end{center}
    \label{tab:lpqWinSize}
\end{table}

\begin{table} [htpb!]
    \caption {Best results found in the experiments that evaluated the performance of handcrafted features, extracted by the use of different texture descriptors.}
    \begin{center}
        \begin{tabular}{c c c} \hline
            Dataset & Texture Descriptor & F-Measure\\\hline
                   & LPQ$_{13}$     & 0.8014\\
            AIA    & LPQ$_{11}$     &\textbf{0.8110}\\
                   & RLBP$_{8,2}$   & 0.8062\\
                   \hline
                   & LPQ$_{13}$     & \textbf{0.9712}\\
            TW     & LPQ$_{11}$     & 0.9567\\
                   & LPQ$_5$        & 0.9591\\
                   \hline
                   & LPQ$_{13}$     & 0.8636\\
            D      & LPQ$_{11}$     & 0.8690\\
                   & LPQ$_9$        & \textbf{0.8773}\\
                   \hline
        \end{tabular}
    \end{center}
    \label{tab:bestOfHcf}
\end{table}
    
In Table \ref{tab:bestOfHcf} it is possible to observe F-measure values of 0.8110, 0.9712 and 0.8773 respectively for the AIA, TW, and D datasets. All of the recently mentioned values were reached by classifying handcrafted features extracted by the LPQ texture descriptor.

The results for the experiments that classified the image samples converted to the grayscale are in Table \ref{tab:grayscale}. In this case, one may observe an improvement in the classification rates in two of the three approached disease groups, being them: AIA and D. In both of them, the F-Measure values reached an average improvement of 2 percentage points, presenting 0.8325 and 0.8907 respectively. For the TW disease group, the greatest found F-Measure value was equivalent to 0.9471, being 0.0241 inferior to the highest value found in past experiments. 
    
Based on these values, it is possible to observe that the best classification rates found in this work until this point, were reached with the LPQ texture descriptor for all the evaluated diseases. With the window size parameter being equivalent to 11 for AIA, and 13 for TW and D.
        
\begin{table} [htpb!]
    \caption {Classification rates found by the tests where handcrafted features were extracted from the image samples converted to the grayscale.}
    \begin{center}
        \begin{tabular}{c c c} \hline
                Dataset & Texture Descriptor & F-Measure\\\hline
                       & LPQ$_{13}$   & 0.8038\\
                AIA    & LPQ$_{11}$   & \textbf{0.8325}\\
                       & RLBP$_{8,2}$ & 0.7775\\
                       \hline
                       & LPQ$_{13}$ & \textbf{0.9471}\\
                TW     & LPQ$_{11}$ & 0.9254\\
                       & LPQ$_5$    & 0.9303\\
                       \hline
                       & LPQ$_{13}$ & \textbf{0.8907}\\
                D      & LPQ$_{11}$ & 0.8851\\
                       & LPQ$_9$    & 0.8696\\
                       \hline
        \end{tabular}
    \end{center}
    \label{tab:grayscale}
\end{table}

The results of the experiments that pseudo-colored the image samples are presented in Table \ref{tab:HSV}. When analyzing them, it can be observed that for the D dataset, a slight difference of 0.006 can be noticed, when compared to the best one found, described in Table \ref{tab:grayscale}. 

\begin{table}[htpb!]
    \caption{Results obtained from the tests where handcrafted features were extracted from image samples pseudo-colored with the HSV colormap.}
    \begin{center}
        \begin{tabular}{c c c} \hline
                Dataset & Texture Descriptor & F-Measure\\\hline
                       & LPQ$_{13}$     & 0.7990\\
                AIA    & LPQ$_{11}$     & \textbf{0.8134}\\
                       & RLBP$_{8,2}$   & 0.7942\\
                       \hline
                       & LPQ$_{13}$     & \textbf{0.9495}\\
                TW     & LPQ$_{11}$     & 0.9327\\
                       & LPQ$_5$        & 0.9183\\
                       \hline
                       & LPQ$_{13}$     & \textbf{0.8967}\\
                D      & LPQ$_{11}$     & 0.8811\\
                       & LPQ$_9$        & 0.8927\\
                       \hline
        \end{tabular}
    \end{center}
    \label{tab:HSV}
\end{table}
    
\begin{figure}[htpb!]
    \centering
    \includegraphics[width=.7\textwidth, keepaspectratio]{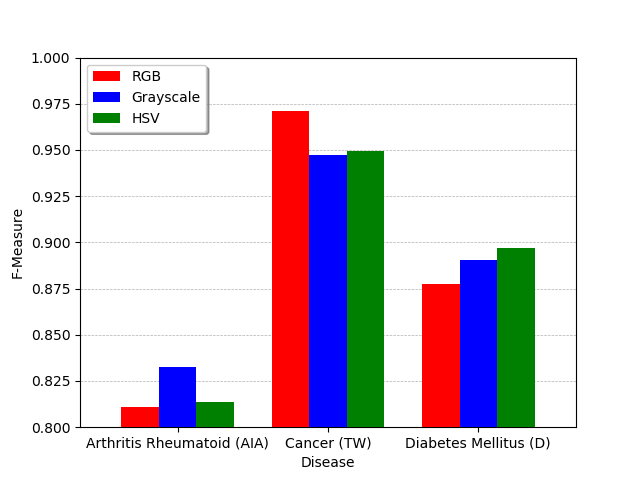}
    \caption{Comparison between the best results obtained from the classifications that aimed to make variations in the image samples' coloration.}
    \label{fig:resul_cores}
\end{figure}

By the end of the experiments that varied the image samples' coloration, it is possible to compare the variations applied with the image samples' on the original color system (RGB) and analyze the overall performance of them. Figure \ref{fig:resul_cores} presents a visual comparison between all of the best results obtained for each of the color variations applied in the image samples. It is possible to observe that a different variation of the image sample was used to achieve its best F-Measure value for each disease group.
        
For the AIA disease group, the image samples that were converted to grayscale reached a better F-Measure than other approaches. Its value was equivalent to 0.8325, nearly 2 percentage points more than the remaining color systems. To achieve this value, the $LPQ_{11}$ texture descriptor was used. 
        
The TW disease group, by its original color system (RGB) presented a performance of approximately 2.5 percentage points greater when compared to the other two. The best F-Measure was equivalent to 0.9712 and it was obtained by the classification of handcrafted features extracted using the $LPQ_{13}$ texture descriptor.
        
Finally, the D disease group achieved its best F-Measure value, equivalent to 0.8967, by the extraction of handcrafted features through the use of the $LPQ_{13}$ texture descriptor and using the image samples pseudo-colored to the HSV colormap.
    
\begin{figure}[htpb!]
    \centering
    \includegraphics[width=.7\textwidth, keepaspectratio]{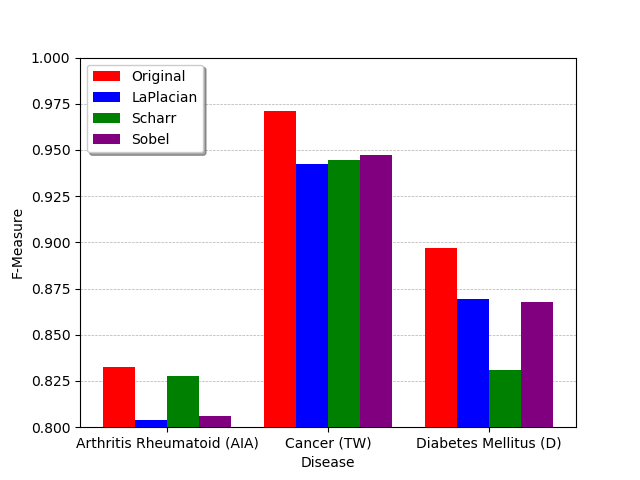}
    \caption{Comparison between the resulting F-Measure values obtained from the classifications where the image samples had their shapes/edges highlighted and the ones obtained from the unchanged image samples (original).}
    \label{fig:resul_edges}
\end{figure}

The F-Measure values obtained from the experiments that highlighted the image samples' edges, are shown in Table \ref{tab:edgesResults}. Figure \ref{fig:resul_edges} presents a graphical comparison between the best results found in this tests and the ones previously obtained.
        
\begin{table}[htpb!]
    \caption{F-Measure values found from the experiment that highlighted the edges of the image samples, through the use of the Laplacian, Sobel and Scharr filters.}
    \begin{center}
    \begin{tabular}{c c c c c} \hline
        \multirow{2}{*}{Disease} & \multirow{2}{*}{Texture Descriptor} & \multicolumn{3}{c}{Filter}\\\cline{3-5}
             &  & Laplacian & Scharr & Sobel\\\hline
                       & LBP$_{8,2}$   & 0.7942 & \textbf{0.8277} & 0.8062\\
                AIA    & LPQ$_{11}$    & 0.7990 & 0.8158 & 0.8062\\
                       & RLBP$_{8,2}$  & 0.8038 & 0.8253 & 0.7918\\
                       \hline
                       & LBP$_{8,2}$   & 0.8992 & 0.9327 & 0.9423\\
                TW     & LPQ$_{13}$    & 0.9424 & 0.9448 & 0.9423\\
                       & RLBP$_{8,2}$  & 0.9279 & 0.9448 & \textbf{0.9471}\\
                       \hline
                       & LBP$_{8,2}$   & 0.8442 & 0.8105 & 0.8171\\
                D      & LPQ$_{13}$    & \textbf{0.8695} & 0.8304 & 0.8675\\
                       & RLBP$_{8,2}$  & 0.8129 & 0.7991 & 0.8128\\
                       \hline
        \end{tabular}
    \end{center}
    \label{tab:edgesResults}
\end{table}
    
For the AIA dataset, it is possible to observe that the Scharr filter had the best results. By the use of such a filter and the LBP texture descriptor, it was reached a F-Measure value of 0.8277. Compared to the best values found by the use of each filter, the recently mentioned value performed approximately 2 percentage points better. 
        
For the TW dataset, the best F-Measure value represented 0.9471, and it was reached using the RLBP texture descriptor and the Sobel Filter. It is worth mentioning that the values found in the classification rates for this dataset showed a high similarity level. These, with few exceptions, were kept around 0.94.
        
Finally, for the D dataset, the highest F-Measure value obtained was equivalent to 0.8695. It was reached by features extracted with the LPQ texture descriptor and edges highlighted by the Laplacian filter. Comparing the F-Measure values obtained in these tests, it is possible to observe a higher performance when the LPQ texture descriptor was used. That shows its potential in the scenario when the image samples are less affected by blur.
    
The next experiments aimed to evaluate the performance of different classification algorithms. The features used as input in the experiments carried out here were extracted by:
        
\begin{itemize}
    \item \textbf{AIA:} the $LPQ_{11}$ texture descriptor, with the image samples converted to the grayscale;
    \item \textbf{TW:} the $LPQ_{13}$ texture descriptor, with the image samples in their original coloration;
    \item \textbf{D:} the $LPQ_{13}$ texture descriptor, with the image samples pseudo-colored to the HSV colormap.
\end{itemize}

The results obtained from the execution of these tests can be seen in Table \ref{tab:alg_cls}. In this one, it is possible to observe that the RF classification algorithm resulted in better classification rates than the other ones. In contrast, the NB algorithm obtained the lowest classification rates in the executed tests. Compared to the ones obtained by the RF algorithm, the NB's F-Measure values were averagely 0.0837 inferior. Such results may be justified by the algorithm's nature, considering that its performance stands out usually in smaller datasets with categorical features.
        
\begin{table}[htpb!]
    \centering
    \caption{Classification rates obtained from the experiments that tested the execution of different classification algorithms. The results found by the classification using the SVM algorithm are also shown for comparison reasons.}
        
    \begin{tabular}{c c c c | c} \hline
        \multirow{2}{*}{Dataset} & \multicolumn{4}{c}{Classification Algorithm}\\\cline{2-5}
        & k-NN & RF & NB & SVM \\\hline
        AIA & 0.7249 & \textbf{0.7847} & 0.7215 & 0.8325 \\
        TW  & 0.8776 & \textbf{0.8920} & 0.7672 & 0.9712 \\
        D   & 0.8286 & \textbf{0.8325} & 0.7694 & 0.8967 \\
        \hline
    \end{tabular}
    \label{tab:alg_cls}
\end{table}
        
\begin{figure}[ht]
    \centering
    \includegraphics[width=.7\textwidth, keepaspectratio]{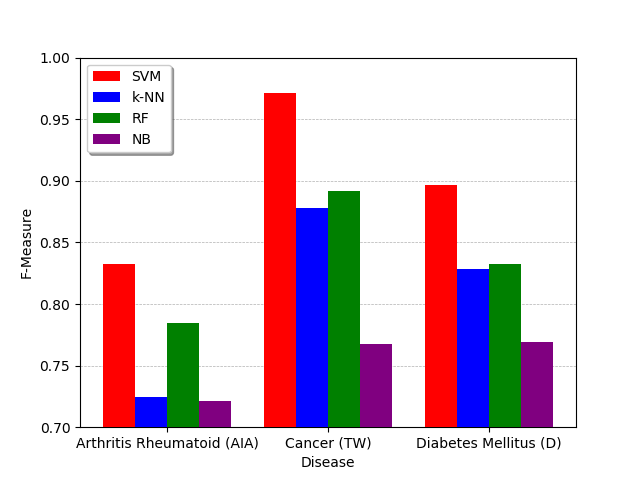}
    \caption{Graphical comparison between the F-Measure values obtained by the classifications performed using different classification algorithms.}
    \label{fig:resul_classif}
\end{figure}
    
A comparison can be seen represented in Figure \ref{fig:resul_classif}. In this one, it is possible to observe the SVM algorithm's superior performance in the same set of features, having its values averagely 0.0638 greater when compared to the values achieved by the RF algorithm, which had the best classification rates in the tests here executed.

\subsubsection{Non-handcrafted modelling}
From this moment, we describe the experiments carried out using CNNs to extract the features and perform the classification for the proposed problem. The best-reached results by each used architecture can be seen in Table \ref{tab:res_cnn}.
    
\begin{table}[htpb!]
    \centering
    \caption{Best results obtained for each CNN architecture, from the experiments that used them to perform the FL and classification.}
    \begin{tabular}{c c c c c c c} \hline
        Dataset & CNN Architecture & Learning Rate & F-Measure\\\hline
               & LeNet5  & 10$^{-3}$ & \textbf{0.7224}\\
        AIA    & MaxNet  & 10$^{-4}$ & 0.6904\\
               & AlexNet & 10$^{-5}$ & 0.6984\\
               \hline
               & LeNet5  & 10$^{-3}$ & 0.8035\\
        TW     & MaxNet  & 10$^{-3}$ & 0.8199\\
               & AlexNet & 10$^{-4}$ & \textbf{0.8721}\\
               \hline
               & LeNet5  & 10$^{-3}$ & 0.6772\\
        D      & MaxNet  & 10$^{-4}$ & 0.7216\\
               & AlexNet & 10$^{-5}$ & \textbf{0.7475}\\
               \hline
    \end{tabular}
    \label{tab:res_cnn}
    
\end{table}

In  Table \ref{tab:res_cnn}, it is possible to observe that for the set of image samples of the AIA disease group, the LeNet5 CNN architecture presented the best F-Measure value. This leads us to conclude that a simpler architecture would be satisfactory for this work's goal in this scenario. The TW and D disease groups' set of image samples otherwise, had their best F-Measure values found by using the AlexNet architecture. This one may be considered as a more robust architecture. However, when these results are compared to the ones found through handcrafted features, it is possible to observe lower classification rates, having a difference in the F-Measure values of 0.1194 (average).

The best results from executing the experiments that employed the Transfer Learning technique may be found in Table \ref{tab:res_tl}. We may notice that the InceptionResNetV2 architecture achieved the worst F-Measure values, around 22 (average) percentage points inferior to the best values found in these experiments.

\begin{table}[htpb!]
    \centering
    \caption{Best classification rates found from executing the tests using the Transfer Learning method, for each evaluated CNN architecture.}
    
    \begin{tabular}{c c c c c c c} \hline
        Dataset & CNN & $n$ & F-Measure\\\hline
               & VGG16             & 4096 & 0.8421\\
        AIA    & InceptionV3       & 4096 & \textbf{0.8468}\\
               & InceptionResNetV2 & 1024 & 0.6579\\
               \hline
               & VGG16             & 2048 & \textbf{0.9327}\\
        TW     & InceptionV3       & 4096 & 0.8823\\
               & InceptionResNetV2 & 2048 & 0.7444\\
               \hline
               & VGG16             & 4096 & \textbf{0.9043}\\
        D      & InceptionV3       & 4096 & 0.7417\\
               & InceptionResNetV2 & 2048 & 0.5982\\
               \hline
    \end{tabular}
    \label{tab:res_tl}

\end{table}
        
For the AIA disease group, the best results were reached using the InceptionV3 architecture and 4096 features. In this case, the F-Measure value was 0.8468. The TW and D disease groups reached their best results using the VGG16 architecture, with the feature quantity reduced to 2048 (TW) and 4096 (D). It is worth mentioning that the best F-Measure values for the AIA and D datasets surpassed the best ones found until this point, obtained using handcrafted-features, by 0.0143 and 0.0065 respectively.

\subsubsection{Classifiers Combination}
Finally, this last experiment evaluates the impacts of combining classifiers.  For this purpose, we selected the most promising classifiers presented until this point.  Table \ref{tab:clasf_comb} briefly introduces them.

\begin{table}[htpb!]
    \caption{Description of the classifiers used to execute the experiments that employed classifiers combination techniques.}
    \begin{center}
    \resizebox{\textwidth}{!}{
        \begin{tabular}{c c c c l c}\hline
            Dataset & ID & Feature Extractor & Classification & Observations & F-Measure\\
                    \hline
                    & 1 & LPQ$_{11}$   & SVM     & Samples converted to the greyscale & 0.8325\\
                    & 2 & LPQ$_{11}$   & RF      & Samples converted to the greyscale & 0.7847\\
            AIA     & 3 & LBP$_{8,2}$  & SVM     & Border enhancement filter = \textit{Scharr} & 0.8277\\
                    & 4 & InceptionV3  & SVM     & \# of \textit{features} = 4096 & 0.8468\\
                    & 5 & VGG16        & SVM     & \# of \textit{features} = 4096 & 0.8421\\
                    & 6 & LeNet5       & LeNet5  & \textit{Learning Rate} = 10$^{-3}$ & 0.7224\\
                    \hline
                    & 1 & LPQ$_{13}$   & SVM     & - & 0.9712\\
                    & 2 & LPQ$_{13}$   & RF      & - & 0.8920\\
            TW      & 3 & RLBP$_{8,2}$ & SVM     & Border enhancement filter = \textit{Sobel} & 0.9471\\
                    & 4 & VGG16        & SVM     & \# of \textit{features} = 2048 & 0.9327\\
                    & 5 & AlexNet      & AlexNet & \textit{Learning Rate} = 10$^{-4}$ & 0.8721\\
                    & 6 & MaxNet       & MaxNet  & \textit{Learning Rate} = 10$^{-3}$ & 0.8199\\
                    \hline
                    & 1 & LPQ$_{13}$   & SVM     & Pseudo-coloring = HSV & 0.8967\\
                    & 2 & LPQ$_{13}$   & RF      & Pseudo-coloring = HSV & 0.8325\\
            D       & 3 & LPQ$_{13}$   & SVM     & Border enhancement filter = \textit{Laplacian} & 0.8695\\
                    & 4 & VGG16        & SVM     & \# of \textit{features} = 4096 & 0.9043\\
                    & 5 & AlexNet      & AlexNet & \textit{Learning Rate} = 10$^{-5}$ & 0.7475\\
                    & 6 & MaxNet       & MaxNet  & \textit{Learning Rate} = 10$^{-4}$ & 0.7216\\
                   \hline
        \end{tabular}
    }
    \end{center}
    \label{tab:clasf_comb}
\end{table}
    
The best results obtained by the combination is reported in Table \ref{tab:resul_comb}.
    
\begin{table}[htpb!]
    \caption{Best F-Measure values found from the experiments that combined classifiers.}
    \centering
    
        \begin{tabular}{c c c c c}\hline
            Dataset & Classifiers (IDs) & Type(s)$^*$ & Rule & F-Measure\\\hline
                    & 3, 4       & N and H & Max & 0.8863\\
                    & 3, 4, 5    & N and H & Max & \textbf{0.8930}\\
            AIA     & 3, 4, 5    & N and H & Sum & 0.8899\\
                    & 1, 3, 4, 5 & N and H & Max & 0.8858\\
                    & 1, 3, 4, 6 & N and H & Sum & 0.8806\\
                    \hline
                    & 1, 2, 5       & N and H & Sum     & 0.9818\\
                    & 1, 3, 4       & N and H & Sum     & 0.9767\\
            TW      & 1, 4, 5       & N and H & Sum     & 0.9789\\
                    & 1, 2, 3, 6    & N and H & Product & 0.9739\\
                    & 1, 2, 3, 5, 6 & N and H & Sum     & \textbf{0.9845}\\
                    \hline
                    & 1, 2       & H        & Max     & 0.9413\\
                    & 1, 2, 4    & N and H  & Sum     & \textbf{0.9513}\\
            D       & 1, 2, 4    & N and H  & Product & 0.9495\\
                    & 1, 2, 4, 5 & N and H  & Sum     & 0.9459\\
                    & 1, 2, 4, 6 & N and H  & Sum     & 0.9445\\
                    \hline
            \multicolumn{5}{l}{$^*$ Type(s) of classifiers involved in the combination.}\\
            \multicolumn{5}{l}{$^{**}$ N stands for ``non-handcrafted'' features, and H stands for ``handcrafted'' features.}    
        \end{tabular}
    
    \label{tab:resul_comb}
\end{table}

For the AIA dataset,  the combination of three classifiers, through the max rule, produced a F-measure value of 0.893, which is 4.62 percentage points better than the best one found so far (available in Table \ref{tab:res_tl}). Among the classifiers used, one used handcrafted features, while the other two used non-handcrafted features. The classifiers with IDs equivalent to 3 and 4 were extremely influential to the classification rates obtained in these tests. Both of them are present in all of the best results found.
    
When analyzing the TW dataset classification rates, it is possible to observe that the best F-Measure value found was equivalent to 0.9845. Approximately 1.33 percentage points more than the best F-Measure value found until this point (observable in Table \ref{tab:lpqWinSize}) for such dataset. In total, five classifiers were combined by the sum rule to achieve these values. The combination counted with three classifiers that used handcrafted features and two that used non-handcrafted features.
    
For the D dataset, the best F-measure value was  0.9513, which is 5.13 percentage points better than the best value found for this dataset (available in Table \ref{tab:res_tl}). This result was possible by combining three classifiers using the sum rule. In this case, two classifiers used handcrafted features and one non-handcrafted. It is worth mentioning that the classifiers with ID 1 and 2, can be considered essential to obtaining these performances. 
    
\begin{figure}[htpb!]
    \centering 
    \begin{subfigure}[b]{\textwidth}
        \centering 
         \includegraphics[width=0.9\textwidth]{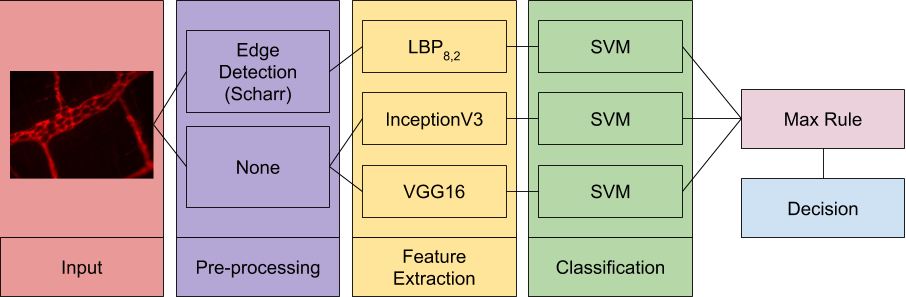}
         \caption{AIA dataset}
         \label{fig:arq_final_aia}
    \end{subfigure}
    \begin{subfigure}[b]{\textwidth}
        \centering 
         \includegraphics[width=0.9\textwidth]{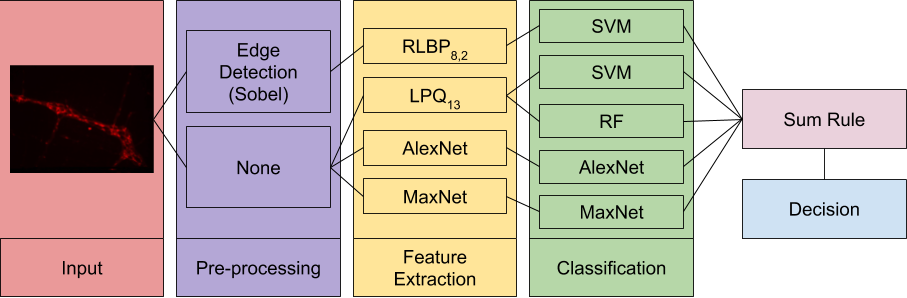}
         \caption{TW dataset}
         \label{fig:arq_final_tw}
    \end{subfigure}
    \begin{subfigure}[b]{\textwidth}
        \centering 
         \includegraphics[width=0.9\textwidth]{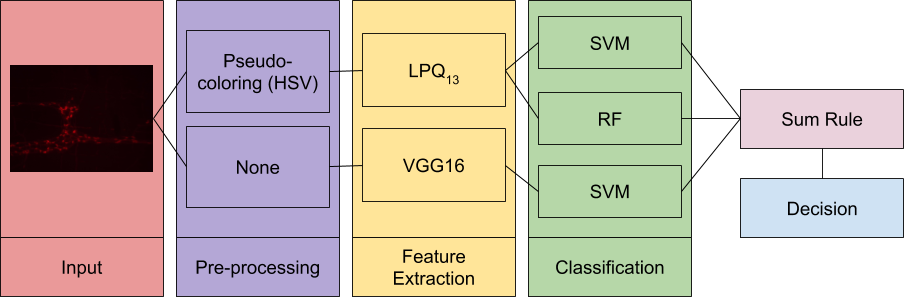}
         \caption{D dataset}
         \label{fig:arq_final_d}
    \end{subfigure}
    \caption{Graphical representation of the configurations architected, so the best classification rates of this work could be achieved, by combining classifiers.}
    \label{fig:arq_final}
\end{figure}

Figure \ref{fig:arq_final} shows a graphical representation of the configurations used to combine the classifiers that achieved the best classification rates presented in this Section, recently displayed in Table \ref{tab:resul_comb}, for each dataset approached in this work.
    
\section{Discussions}
    \label{sec:discussions}

In this section, we discuss the results presented so far. It will be conducted by looking for answers to the following questions, aiming to get a more comprehensive understanding of the behavior of the methods and techniques experimented on the evaluated classification tasks:

\begin{itemize}

    \item Which features provided the best results in each feature representation scenario and disease group?
    
    \item Which feature representation provided the best results?
    
    \item Which disease group is easier/harder to predict?
    
    \item Have the fusion strategies contributed to improving the results?
    
    
\end{itemize}

\subsection{Which features provided the best results in each feature representation scenario and disease group?}

This question is three-fold, i.e., we have to consider the three different feature representation scenarios used in each dataset: The handcrafted features (LBP, RLBP, and LPQ); The non-handcrafted features with pre-configured CNNs (MaxNet, LeNet5, and AlexNet); and the non-handcrafted features with transfer learning (VGG16, InceptionV3, and InceptionResNetV2).

To answer the question, i.e., define the best features in each representation scenario/disease group, we used the statistical evaluation protocol proposed in Charte et al. (2015) \cite{charte2015mlsmote}. Using this protocol, we calculate the ranking of the f-measures classification results obtained in all experiments with the different features based on the Friedman statistical test. The classification performances using the three different types of feature representation are ranked per disease group (from first to last), and an average rank is calculated. Then a general average ranking is computed for each one of the three feature representation scenarios.

The average rankings obtained with the previously described statistical test for the handcrafted features are presented in Table \ref{fried-rank-hand}. We may observe that for the AIA disease, the best-ranked feature with an average ranking of 1.50 was LBP, while for the TW and D diseases, the best-ranked feature was LPQ, with an average ranking of 1.00. Moreover, considering all disease groups, the best ranked handcrafted feature is LPQ, with an average of 1.44.

\begin{table}[htbp]
\centering
\caption{Average ranking of the classification results for the handcrafted features.}
\begin{tabular}{rcccc}
\cline{2-5}
\multicolumn{1}{l}{} & \multicolumn{3}{c}{Disease} & \multirow{2}{*}{\begin{tabular}[c]{@{}c@{}}Overall\\ Avg. Ranking\end{tabular}} \\ \cline{2-4}
\multicolumn{1}{l}{} & AIA & TW & D &  \\ \hline
LBP & 1.50 & 2.50 & 2.33 & 2.11 \\
RLBP & 2.00 & 2.17 & 2.67 & 2.28 \\
LPQ & 2.33 & 1.00 & 1.00 & 1.44 \\ \hline
\end{tabular}
\label{fried-rank-hand}
\end{table}

Table \ref{fried-rank-pre} shows the average rankings for the non-handcrafted features obtained with pre-configured CNNs. We may observe that in all disease groups, AlexNet obtained the best classification results, reaching an average ranking of 1.50 for the AIA disease and 1.00 for TW and D diseases. Therefore, AlexNet is also the best-ranked feature in general, i.e., considering the three disease groups.

\begin{table}[htbp]
\centering
\caption{Average ranking of the classification results for the non-handcrafted features obtained with pre-configured CNNs.}
\begin{tabular}{rcccc}
\cline{2-5}
\multicolumn{1}{l}{} & \multicolumn{3}{c}{Disease} & \multirow{2}{*}{\begin{tabular}[c]{@{}c@{}}Overall\\ Avg. Ranking\end{tabular}} \\ \cline{2-4}
\multicolumn{1}{l}{} & AIA & TW & D &  \\ \hline
MaxNet & 2.50 & 2.00 & 2.00 & 2.17 \\
LeNet5 & 2.00 & 3.00 & 3.00 & 2.67 \\
AlexNet & 1.50 & 1.00 & 1.00 & 1.17 \\ \hline
\end{tabular}
\label{fried-rank-pre}
\end{table}

Table \ref{fried-rank-tl} presents the average ranking for the non-handcrafted features obtained with transfer learning. We may note that VGG16 achieved the best overall average ranking considering all disease groups (1.13), with an average ranking of 1.40 for AIA and 1.00 for both TW and D disease groups.

\begin{table}[htbp]
\centering
\caption{Average ranking of the classification results for the non-handcrafted features obtained with transfer learning.}
\begin{tabular}{rcccc}
\cline{2-5}
\multicolumn{1}{l}{} & \multicolumn{3}{c}{Disease} & \multirow{2}{*}{\begin{tabular}[c]{@{}c@{}}Overall\\ Avg. Ranking\end{tabular}} \\ \cline{2-4}
\multicolumn{1}{l}{} & AIA & TW & D &  \\ \hline
VGG16 & 1.40 & 1.00 & 1.00 & 1.13 \\
InceptionV3 & 1.60 & 2.00 & 2.00 & 1.87 \\
InceptionResNetV2 & 3.00 & 3.00 & 3.00 & 3.00 \\ \hline
\end{tabular}
\label{fried-rank-tl}
\end{table}

\subsection{Which feature representation provided the best results?}

To answer the second question with statistical significance, we applied the Wilcoxon Statistical Test. We hypothesize that the top-10 f-measures obtained with the handcrafted features are higher than the top-10 f-measures obtained with the non-handcrafted features (considering both pre-configured CNNs and transfer learning scenarios). The test was computed three times, one for each disease group.

The \textit{z-scores} and \textit{p-values} obtained with the Wilcoxon Statistical Test are reported in Table \ref{wilcoxon-feat-rep}. Considering a threshold of 0.05, we can statistically affirm that for TW and D disease groups, the handcrafted features obtained better classification results than the non-handcrafted features, since their \textit{p-values} are below the threshold with 0.0025 and 0.0035 values, respectively. However, considering the same threshold, we cannot statistically confirm that the handcrafted features are better than the non-handcrafted for the AIA disease, since its \textit{p-value} resulted in 0.1931.

\begin{table}[htbp]
\centering
\caption{Wilcoxon statistical tests for the top-10 classification results with handcrafted features versus non-handcrafted.}
\begin{tabular}{rcc}
\hline
\multicolumn{1}{c}{Disease} & \textit{z-score} & \textit{p-value} \\ \hline
AIA & 19 & 0.1931 \\
TW & 0 & 0.0025 \\
D & 1 & 0.0035 \\ \hline
\end{tabular}
\label{wilcoxon-feat-rep}
\end{table}

\subsection{Which disease group is easier/harder to predict?}

To answer the third question, we have also used the Wilcoxon Statistical Test. We have applied the test hypothesizing that the f-measures classification results obtained in all experiments for a certain disease group are lower than the f-measures obtained in another disease group. We have computed the test three times since we have three different disease groups. The \textit{z-scores} and \textit{p-values} obtained with the Wilcoxon Statistical Test for this scenario are reported in Table \ref{wilcoxon-groups}. 

\begin{table}[htbp]
\centering
\caption{Wilcoxon statistical tests comparing the classification results obtained in the different diseases groups.}
\begin{tabular}{rcl}
\cline{2-3}
\multicolumn{1}{l}{} & \textit{z-score} & \textit{p-value} \\ \hline
AIA versus TW & 143 & $6.5 \times e^{-12}$ \\
D versus TW & 210 & $7.0 \times e^{-11}$ \\
AIA versus D & 820 & $7.0 \times e^{-4}$ \\ \hline
\end{tabular}
\label{wilcoxon-groups}
\end{table}

Considering a threshold of 0.05, we can statistically affirm that the classification results obtained in AIA disease are lower than the results obtained in TW disease, since the test achieved a \textit{p-value} of $6.5 \times e^{-12}$ (below the threshold). We can also affirm that the results obtained in D disease are lower than the results obtained in TW, since the test obtained a \textit{p-value} of $7.0 \times e^{-11}$. Thus, we can conclude that TW disease is the ``easiest'' one to predict, as it achieved classification results higher than the two other disease groups, i.e., AIA and D.

Furthermore, we can also observe in Table \ref{wilcoxon-groups} that the classification results obtained in AIA disease are lower then the ones obtained in D disease group, since the test achieved a \textit{p-value} of $7.0 \times e^{-4}$ (below the threshold). Thus, we can conclude that AIA disease is the ``hardest'' one to predict, as it achieved classification results lower than the two other disease groups (D and TW).

\subsection{Have the fusion strategies contributed to improve the results?}

To answer this question, our hypothesis is that the top-5 classification results with the combinations of the classifiers' outputs are higher than the top-5 classification results without the combination of outputs in each disease group. The \textit{z-scores} and \textit{p-values} obtained with the Wilcoxon Statistical Test for this scenario are reported in Table \ref{wilcoxon-fusion}. 

Considering a threshold of 0.05, we can statistically affirm that the fusion strategies did contribute to improving the overall classification results, since the \textit{p-values} of the Wilcoxon test achieved 0.0216 for all disease groups.

\begin{table}[htbp]
\centering
\caption{Wilcoxon statistical tests comparing the classification results before and after combining the classifiers outputs.}
\begin{tabular}{rcc}
\cline{2-3}
\multicolumn{1}{l}{} & \textit{z-score} & \textit{p-value} \\ \hline
AIA & 0.0 & 0.0216 \\
TW & 0.0 & 0.0216 \\
D & 0.0 & 0.0216 \\ \hline
\end{tabular}
\label{wilcoxon-fusion}
\end{table}

\section{Concluding remarks and future works}
    \label{sec:conclusions}

Nowadays, as the pre-clinical research advances, it is created the tendency of building methods capable of supporting such activities, aiming to decrease the manual work demanded and automatically analyze the conceived data. Thus, we proposed a method to automatically identifying chronic degenerative diseases using EGC image samples.  Two types of features were considered (handcrafted and non-handcrafted). Both methodologies had their efficiency measured. Furthermore, a hybrid methodology was evaluated. Classifiers formed from such methodologies were combined to ascertain if it is possible to take advantage of a potential complementarity between them.

By experiments that were performed using handcrafted features, it was possible to observe a tendency of better results when the LPQ texture descriptor and the SVM classification algorithm were used. This descriptor's efficiency can be justified by the fact that it was developed aiming to be efficient for blurred images, like the ones existing in this work's dataset. At the end of these experiments' execution, we may observe that the best F-measure values founded reached 0.8325, 0.9712 and 0.8967 for AIA, TW, and D datasets, respectively. It is worth mentioning that for the AIA and D image samples, a pre-processing was applied, in which these were converted to grayscale (AIA) and pseudo-colored to the HSV colormap (D).

In the second approach used, non-handcrafted features were extracted using Feature Learning methods. Experiments were accomplished using CNN architectures and presented regular classification rates, but inferior to those found until that point. Better classification rates were generated by using a Transfer Learning strategy. The tests that led such results, extracted features using CNN architectures pre-trained with the ImageNet dataset, performing the classification later with the SVM classification algorithm. The best F-measure values found at the end of these tests were 0.8468, 0.9327, and 0.9043 for AIA, TW, and D datasets, respectively. Therefore, a slight improvement may be observed compared to the best ones, found by the classification of handcrafted features, for the AIA and D datasets.

Lastly, classifiers from both approaches were combined. Different configurations/architectures were created, and the target classifiers were combined by the sum, max, and product rules. By the use of this approach, this work's best results were obtained. The final F-Measure values were equal to 0.8930, 0.9845, and 0.9513 to the AIA, TW, and D datasets, respectively. These results were obtained by combining respectively three, five, and three classifiers. In all of them, classifiers created using both handcrafted and non-handcrafted features. 

At the end of this work, it was possible to conclude that the classification of EGC images, aiming to identify a target disease's presence in the data samples, presented great results by classifying features extracted both on handcrafted and non-handcrafted modes. However, combining both strategies achieved the best performances for all disease groups. This corroborates the effectiveness of the proposed method.

In future works, we intend to expand the dataset aiming to include other types of cells, such as the Enteric Neuron. It could be useful to create an alternative understanding of the studied diseases and evaluate if it can improve the results described in this work. 

We also intend to use the data samples as input in a segmentation methodology, using the isolated EGC to create a morphometric and quantitative analysis, aiming to perform the classification more precisely. We plan to adapt our method to analyze samples of individuals under different treatments so that it can i) automatically identify the proximity level to both classes approached in this work (control/healthy and unhealthy); ii) rank the performed treatments, displaying the most successful ones. Finally, we plan to: i) test new data augmentation methods; ii) crate a dedicated CNN architecture; iii) evaluate other texture descriptors; and iv) test different feature selection methods.

\section*{Acknowledgements}
We thank the Brazilian Research Support Agency CNPq - National Council for Scientific and Technological Development.

\section*{Conflict of interest}
The authors declare that they have no conflict of interest.

\bibliographystyle{unsrt}  
\bibliography{main}  

\begin{thebibliography}{10}

\bibitem{furness06}
J.~B. Furness.
\newblock The enteric nervous system.
\newblock {\em Blackwell Publishing}, page 290, 2006.

\bibitem{Sharkey2015}
Keith~A. Sharkey.
\newblock Emerging roles for enteric glia in gastrointestinal disorders.
\newblock {\em The Journal of clinical investigation}, 125, 2015.

\bibitem{Sharkey2012}
Brian Gulbransen and Keith Sharkey.
\newblock Novel functional roles for enteric glia in the gastrointestinal
  tract.
\newblock {\em Nature reviews. Gastroenterology \& hepatology}, 9, 08 2012.

\bibitem{ruhl11}
A.~Ruhl.
\newblock Glial cells in the gut.
\newblock {\em Neurogastroenterology and Motility}, 17:777 -- 790, 2005.

\bibitem{degiorgio12}
R.~De Giorgio et~al.
\newblock Enteric glia and neuroprotection: basic and clinical aspects.
\newblock {\em American Journal of Physiology: Gastrointestinal and Liver
  Physiology}, 303:G887--G893, 2012.

\bibitem{Piovezana2019}
G.~D.~Piovezana Bossolani et~al.
\newblock Rheumatoid arthritis induces enteric neurodegeneration and jejunal
  inflammation, and quercetin promotes neuroprotective and anti-inflammatory
  actions.
\newblock {\em Life Sciences}, December 2019.

\bibitem{Panizzon2019}
C.~P. N.~B. Panizzon et~al.
\newblock Ethyl acetate fraction from trichilia catigua confers partial
  neuroprotection in components of the enteric innervation of the jejunum in
  diabetic rats.
\newblock {\em Cellular Physiology and Biochemistry}, 53:76--86, 2019.

\bibitem{Panizzon2016}
C.~P. N.~B. Panizzon et~al.
\newblock Desired and side effects of the supplementation with l-glutamine and
  l-glutathione in enteric glia of diabetic rats.
\newblock {\em Acta Histochemica}, 118:625--631, July 2016.

\bibitem{lbp}
T.~Ojala, M.~Pietik\"ainen, and T.~M\"aenp\"a\"a.
\newblock Multiresolution gray-scale and rotation invariant texture
  classification with local binary patterns.
\newblock {\em IEEE Transactions on Pattern Analysis and Machine Intelligence},
  24(7):971--987, 2002.

\bibitem{zhao2013completed}
Yang Zhao, Wei Jia, Rong-Xiang Hu, and Hai Min.
\newblock Completed robust local binary pattern for texture classification.
\newblock {\em Neurocomputing}, 106:68--76, 2013.

\bibitem{lpq}
V.~Ojansivu and J.~Heikkil\"a.
\newblock Blur insensitive texture classification using local phase
  quantization.
\newblock {\em Image and Signal Processing}, pages 236--243, 2008.

\bibitem{lenet5-89}
Yann Lecun.
\newblock Generalization and network design strategies.
\newblock In R.~Pfeifer, Z.~Schreter, F.~Fogelman, and L.~Steels, editors, {\em
  Connectionism in perspective}, page~20. Elsevier, 1989.

\bibitem{alexnet12}
Alex Krizhevsky, Ilya Sutskever, and Geoffrey~E Hinton.
\newblock Imagenet classification with deep convolutional neural networks.
\newblock In F.~Pereira, C.~J.~C. Burges, L.~Bottou, and K.~Q. Weinberger,
  editors, {\em Advances in Neural Information Processing Systems}, pages
  1097--1105. Curran Associates, Inc., 2012.

\bibitem{roecker18}
M.~N. Roecker, Y.~M.~G. Costa, J.~L.~R. Almeida, and G.~Matsushita.
\newblock Automatic vehicle type classification with convolutional neural
  networks.
\newblock In {\em Proceedings of the International Conference on Systems,
  Signals and Image Processing}, pages 1--5, 06 2018.

\bibitem{vgg16}
Karen Simonyan and Andrew Zisserman.
\newblock Very deep convolutional networks for large-scale image recognition.
\newblock {\em Computing Research Repository (CoRR)}, abs/1409.1556, 2015.

\bibitem{inceptionv3}
Christian Szegedy, Vincent Vanhoucke, Sergey Ioffe, Jonathon Shlens, and
  Zbigniew Wojna.
\newblock Rethinking the inception architecture for computer vision.
\newblock {\em Computing Research Repository (CoRR)}, abs/1512.00567, 2015.

\bibitem{inceptionresnet}
Christian Szegedy, Sergey Ioffe, and Vincent Vanhoucke.
\newblock Inception-v4, inception-resnet and the impact of residual connections
  on learning.
\newblock {\em Computing Research Repository (CoRR)}, abs/1602.07261, 2016.

\bibitem{imagenet}
J.~Deng, W.~Dong, R.~Socher, L.-J. Li, K.~Li, and L.~Fei-Fei.
\newblock {ImageNet: A Large-Scale Hierarchical Image Database}.
\newblock In {\em CVPR09}, page~8, 2009.

\bibitem{kittler1998}
Josef Kittler, Mohamad Hatef, Robert~PW Duin, and Jiri Matas.
\newblock On combining classifiers.
\newblock {\em IEEE Transactions on Pattern Analysis and Machine Intelligence},
  20(3):226--239, 1998.

\bibitem{maenpaa2003}
T.~I. M\"aenp\"a\"a.
\newblock {\em The Local Binary Pattern Approach to Texture Analysis:
  Extensions and Applications}.
\newblock PhD thesis, University of Oulu, Oulu, 2003.

\bibitem{costa2012music}
Yandre M.~G. Costa, L.~E.~S. Oliveira, Alessandro~L. Koerich, Fabien Gouyon,
  and Jefferson~G. Martins.
\newblock Music genre classification using lbp textural features.
\newblock {\em Signal Processing}, 92(11):2723--2737, 2012.

\bibitem{nanni_ICTAI_2016}
Loris Nanni, Yandre M.~G. Costa, Diego~R. Lucio, Carlos~N. Silla, and Sheryl
  Brahnam.
\newblock Combining visual and acoustic features for bird species
  classification.
\newblock In {\em Proceedings of the IEEE International Conference on Tools
  with Artificial Intelligence}, pages 396--401, 2016.

\bibitem{freitas16}
G.~K. Freitas, Y.~M.~G. Costa, and R.~L. Aguiar.
\newblock Using spectrogram to detect north atlantic right whale calls from
  audio recordings.
\newblock In {\em Proceedings of the International Conference of the Chilean
  Computer Science Society}, pages 1--6, 2016.

\bibitem{felipe2019identification}
Gustavo~Z. Felipe, Rafael~L. Aguiat, Yandre M.~G. Costa, Carlos~N. Silla,
  Sheryl Brahnam, Loris Nanni, and Shannon McMurtrey.
\newblock Identification of infants\textquotesingle cry motivation using
  spectrograms.
\newblock In {\em Proceedings of the International Conference on Systems,
  Signals and Image Processing}, pages 181--186, 2019.

\bibitem{paulino2018brazilian}
Marco Aurelio~D. Paulino, Alceu~S. Britto~Junior, Alisson~R. Svaigen, Linnyer
  B.~R. Aylon, Luiz E.~S. Oliveira, and Yandre M.~G. Costa.
\newblock A brazilian speech database.
\newblock In {\em Proceedings of the IEEE International Conference on Tools
  with Artificial Intelligence}, pages 234--241, 2018.

\bibitem{felipe2017}
G.~Z. Felipe, Y.~M.~G. Costa, and L.~G. Helal.
\newblock Acoustic scene classification using spectrograms.
\newblock In {\em Proceedings of the International Conference of the Chilean
  Computer Science Society}, page~7, 2017.

\bibitem{pereira2020covid}
Rodolfo~M Pereira, Diego Bertolini, Lucas~O Teixeira, Carlos~N Silla~Jr, and
  Yandre M.~G. Costa.
\newblock Covid-19 identification in chest x-ray images on flat and
  hierarchical classification scenarios.
\newblock {\em Computer Methods and Programs in Biomedicine}, 194:105532, 2020.

\bibitem{rlbp}
J.~Chen et~al.
\newblock {RLBP}: Robust local binary pattern, 2013.
\newblock Center for Machine Vision Research, University of Oulu, Oulu.

\bibitem{costa2013}
Y.~M.~G. Costa.
\newblock {\em Reconhecimento de g\^eneros musicais utilizando espectrogramas
  com combina\c{c}\~ao de classificadores}.
\newblock PhD thesis, Federal University of Paran\'a, Curitiba, Brasil, 2013.

\bibitem{norvig}
Stuart~J. Russell and Peter Norvig.
\newblock {\em Artificial Intelligence: A Modern Approach}.
\newblock Pearson Education, 3 edition, 2010.

\bibitem{Duda01}
Richard~O. Duda, Peter~E. Hart, and David~G. Stork.
\newblock {\em Pattern Classification}.
\newblock Wiley, New York, 2 edition, 2001.

\bibitem{tin95}
Tin~Kam Ho.
\newblock Random decision forests.
\newblock In {\em Proceedings of the International Conference on Document
  Analysis and Recognition}, pages 278--282, 1995.

\bibitem{bengio2013}
Y.~Bengio, A.~Courville, and P.~Vincent.
\newblock Representation learning: A review and new perspectives.
\newblock In {\em IEEE Transactions on Pattern Analysis and Machine
  Intelligence}, pages 1798--1828, 2013.

\bibitem{Srivastava14}
N.~Srivastava, G.~Hinton, A.~Krizhevsky, I.~Sutskever, and R.~Salakhutdinov.
\newblock Dropout: A simple way to prevent neural networks from overfitting.
\newblock {\em Journal of Machine Learning Research}, 15:1929--1958, 2014.

\bibitem{Liu2015}
Tianyi Liu, Shuangsang Fang, Yuehui Zhao, and Jing Zhang.
\newblock Implementation of training convolutional neural networks.
\newblock {\em Computing Research Repository (CoRR)}, abs/1506.01195, 2015.

\bibitem{lecun89}
Y.~Lecun et~al.
\newblock Backpropagation applied to handwritten zip code recognition.
\newblock {\em Neural Computation}, 1:541--551, 1989.

\bibitem{vargas16}
A.~C.~G. Vargas and A.~V. C.~N. Paes.
\newblock Um estudo sobre redes neurais convolucionais e sua aplicação em
  detecção de pedestres.
\newblock In {\em Proceedings of the Conference on Graphics, Patterns and
  Images}, page~4, 2016.

\bibitem{matsu2018}
Gustavo H.~G. Matsushita, Adam~H. Sugi, Yandre M.~G. Costa, Alexander Gomez-A,
  Claudio Da~Cunha, and Luiz E.~S. Oliveira.
\newblock Phasic dopamine release identification using convolutional neural
  network.
\newblock {\em Computers in Biology and Medicine}, 114:103466, 2019.

\bibitem{Pan10}
S.~J. Pan and Q.~Yang.
\newblock A survey on transfer learning.
\newblock {\em IEEE Transactions on Knowledge and Data Engineering},
  22(10):1345--1359, Oct 2010.

\bibitem{costa2017evaluation}
Yandre M.~G. Costa, Luiz E.~S. Oliveira, and Carlos~N. Silla~Jr.
\newblock An evaluation of convolutional neural networks for music
  classification using spectrograms.
\newblock {\em Applied Soft Computing}, 52:28--38, 2017.

\bibitem{nanni17}
Loris Nanni, Stefano Ghidoni, and Sheryl Brahnam.
\newblock Handcrafted vs non-handcrafted features for computer vision
  classitcation.
\newblock {\em Pattern Recognition}, 71:158--172, 05 2017.

\bibitem{frez2017}
F.~C.~V. Frez et~al.
\newblock Restoration of density of interstitial cells of cajal in the jejunum
  of diabetic rats after quercetin supplementation.
\newblock {\em Revista Espanola de Enfermedades Digestivas}, 109:190--195,
  2017.

\bibitem{vincentini17}
G.~E. Vicentini et~al.
\newblock Does l-glutamine-supplemented diet extenuate no-mediated damage on
  myenteric plexus of walker 256 tumor-bearing rats?
\newblock {\em Food Research International}, 101:24--34, 2017.

\bibitem{souza11}
I.~D. Souza et~al.
\newblock Analysis of myosin-v immunoreactive myenteric neurons from arthritic
  rats.
\newblock {\em Arquivos de Gastroenterologia}, 48:205--210, 2011.

\bibitem{pereira11}
R.~V. Pereira et~al.
\newblock L-glutamine supplementation prevents myenteric neuron loss and has
  gliatrophic effects in the ileum of diabetic rats.
\newblock {\em Digestive Disease and Science}, 56:3507--3516, 2011.

\bibitem{libsvm}
C.~Chang and C.~Lin.
\newblock Libsvm: A library for support vector machines, 2013.
\newblock National Taiwan University, Taipei.

\bibitem{scikit-learn}
F.~Pedregosa, G.~Varoquaux, A.~Gramfort, V.~Michel, B.~Thirion, O.~Grisel,
  M.~Blondel, P.~Prettenhofer, R.~Weiss, V.~Dubourg, J.~Vanderplas, A.~Passos,
  D.~Cournapeau, M.~Brucher, M.~Perrot, and E.~Duchesnay.
\newblock Scikit-learn: Machine learning in {P}ython.
\newblock {\em Journal of Machine Learning Research}, 12:2825--2830, 2011.

\bibitem{2015keras}
Fran\c{c}ois Chollet et~al.
\newblock Keras.
\newblock \url{https://keras.io}, 2015.

\bibitem{adamOpt}
Diederik~P. Kingma and Jimmy Ba.
\newblock Adam: A method for stochastic optimization.
\newblock In {\em Proceedings of the International Conference for Learning
  Representations}, page~15, 2014.

\bibitem{charte2015mlsmote}
Francisco Charte, Antonio Rivera, Mar{\'\i}a del Jesus, and Francisco Herrera.
\newblock {MLSMOTE}: Approaching imbalanced multilabel learning through
  synthetic instance generation.
\newblock {\em Knowledge-Based Systems}, 89:385--397, 2015.

\end{thebibliography}

\end{document}